\documentclass{optica-article}

\input{opticajournal.sty} 

\makeatletter
\renewenvironment{abstract}%
  {\vskip1pc\noindent\textbf{Abstract:\space}}%
  {\par\vskip12pt}
\makeatother

\articletype{Research Article}

\usepackage{lineno}
\usepackage[utf8]{inputenc}
\usepackage[T1]{fontenc}
\usepackage{siunitx}
\usepackage{hyperref}
\usepackage{placeins}
\usepackage{overpic}
\usepackage{multirow}
\usepackage{booktabs}
\usepackage{colortbl}  
\usepackage{xcolor}    
\usepackage{amsmath}

\usepackage{amssymb}
\usepackage{algorithm}
\usepackage{algpseudocode}

\definecolor{headerblue}{RGB}{41, 82, 135}
\definecolor{lightblue}{RGB}{219, 229, 241}
\definecolor{rowgray}{RGB}{247, 248, 250}

\begin{document}

\title{Label-free cell counting and viability prediction with brightfield imaging and deep learning}

\author{Amir Reza Vazifeh,\authormark{1} Christian Zeigler,\authormark{2} Sornanathan Meyyappan,\authormark{3} Richard Jeske,\authormark{4} Jason W. Fleischer,\authormark{1,*}}

\address{%
  \authormark{1}Department of Electrical and Computer Engineering, Princeton University, Princeton, NJ 08544, USA\\
  \authormark{2}Waters Corporation, Immerse Cambridge, 301 Binney Street, Suite 102, Cambridge, MA 02142, USA \\
  \authormark{3}Waters Corporation, 34 Maple St, Milford, MA 01757, USA \\
  \authormark{4}Waters Corporation, Immerse Delaware, 590 Avenue 1743, Newark, DE 19130, USA
}

\email{\authormark{*}jasonf@princeton.edu}



\begin{abstract}
Cell viability assessment is a core requirement in cell culture systems, with critical applications in biopharmaceutical manufacturing and drug development. Conventionally, it is measured by adding membrane-impermeable dyes to a sample (a process called staining), which allows compromised cell membranes to be distinguished from intact ones. However, staining has several limitations: (a) chemical agents can perturb normal cellular processes of the cells being measured, (b) it is often ambiguous to assign viability to individual cells whose membrane integrity is only partially compromised, (c) photobleaching can undermine measurement accuracy over time when using fluorescent stains, and (d) staining cannot be performed in situ or in real time. Here, we show that (1) stained cells captured under brightfield imaging contain sufficient information to distinguish live and dead cells, and (2) cells captured under unstained brightfield imaging exhibit similar image features to their stained counterparts, enabling models trained on stained cells to generalize to unstained ones. We then report the development and validation of ViabiLens, an AI-assisted software tool for label-free cell viability analysis. ViabiLens combines a cell detection model for localizing individual cells with a convolutional neural network (CNN) classifier for live/dead prediction, paired with an interactive UMAP-based viewer for visualizing and exploring individual cells across the sample. Evaluated on Chinese Hamster Ovary (CHO) cells spanning a wide range of viability conditions, ViabiLens achieves a mean absolute error of 2.33\% on unstained samples against fluorescence-based reference measurements. The demo of ViabiLens software is available at \url{https://amirrezavazifeh.github.io/ViabiLens-Project-Page/}.
\end{abstract}


\section{Introduction}
\label{sec:intro}

Cell culture technology is central to the biopharmaceutical industry, where living cells are grown in bioreactors to produce therapeutic proteins such as monoclonal antibodies \cite{Jyothilekshmi:21}. CHO cells are the mammalian cell line most widely used for this purpose~\cite{Wurm:04, Kim:12}. Biologics produced this way now dominate the pharmaceutical industry, accounting for 8 of the 10 top-selling drugs in 2023 \cite{Verdin:24} and reaching a record 40\% share of FDA approvals in 2022, a steady rise over the past 25 years \cite{Martins:23}. Key to consistent and predictable operation of cell culture systems is the ability to accurately know the total number of cells (cell count) and the percentage of live or dead cells (cell viability). Beyond bio-production, cells themselves are increasingly being used as therapeutics, including CAR-T cells, stem cells, and tumor-infiltrating lymphocyte therapies~\cite{Reddy:20}. As dying cells can reduce therapeutic efficacy and pose immunological risks to patients, accurate measurement of cellular health is particularly important~\cite{Cai:24}.

\begin{figure}[t]
    \centering
    \includegraphics[width=0.95\linewidth]{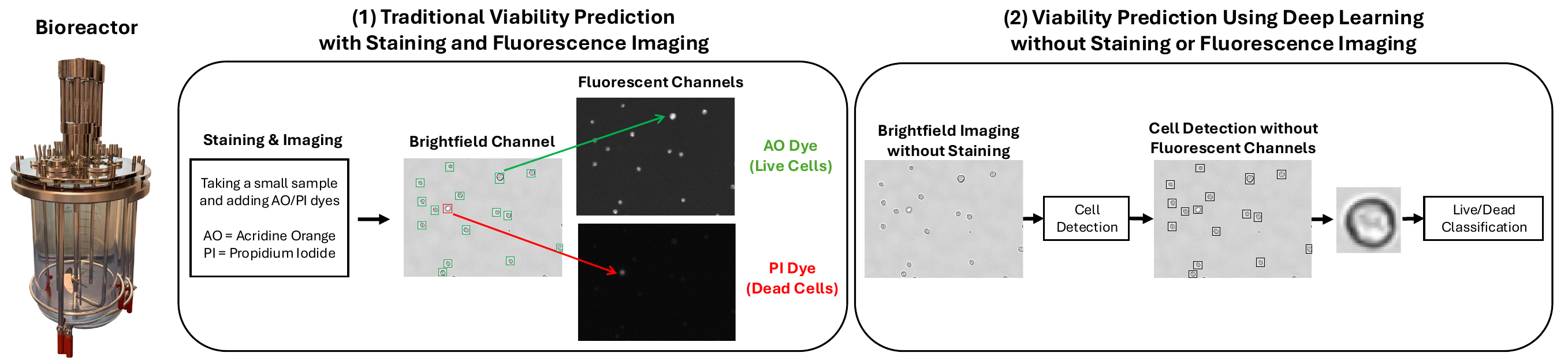}
    \caption{Comparison of traditional and proposed methods for cell viability assessment. Left: A bioreactor used for cell culture. Middle: Traditionally, cells are stained and fluorescently imaged to distinguish live from dead cells. Right: The proposed deep-learning-based method detects and classifies individual cells as live or dead directly from brightfield images, without staining or fluorescence imaging. In either case, sample-level viability is computed as the ratio of live cells to the total cell count.}
    \label{fig:overview}
\end{figure}

Conventional approaches for measuring cell viability typically rely on membrane-impermeable dyes (e.g., Trypan Blue or the combination of acridine orange (AO) and propidium iodide (PI)), where loss of membrane integrity is used as an indicator of cell death~\cite{Naveen:26, Piccinini:17}. In AO/PI assays, live and dead cells emit fluorescence at different wavelengths based on membrane permeability, enabling their discrimination via fluorescence imaging (Fig.~\ref{fig:overview}). However, this approach is subject to several limitations. First, the requirement to add chemical agents introduces inherent disruption to the sample: cells exposed to staining agents may behave differently from unstained cells, and in some cases the staining process itself can kill the cells being measured~\cite{Kwok:04}. Dye exposure can also cause dead cells to rupture and break apart into debris, causing them to be under-counted and biasing measured viability upward~\cite{Chan:15, Chan:20}. Second, the boundary between live and dead populations is not always clear-cut, as  cells sometimes appear in both fluorescent channels simultaneously. In addition, PI positivity does not always mean death. For example, cells stressed by heat  can take up PI immediately, but their membranes often reseal shortly afterwards~\cite{Davey:11}. Third, photobleaching (the progressive loss of fluorescent signal that occurs when a fluorophore is exposed to excitation light~\cite{Lichtman:05}) can reduce measurement reliability over time. Finally, because a physical sample must be pulled manually from the bioreactor and chemically prepared prior to imaging, these measurements cannot be performed in situ or in real time. This manual process is also time-consuming, increases the risk of contaminating the batch, and prevents continuous monitoring of cell health~\cite{Chan:25}.

Recent advances in artificial intelligence have provided an alternative approach to bypass the staining process altogether. A common example is virtual staining, in which neural networks with U-Net-like architectures~\cite{Ronneberger:15} are trained to predict fluorescence images directly from brightfield or phase contrast inputs~\cite{Hu:22, Christiansen:18, Liu:20, Rivenson:19}.  These methods, however, require accurately co-registered image pairs for training, which are not always available in practice or are laborious to acquire at the volumes needed for proper training. Furthermore, virtual staining shares some of the same limitations as physical staining: some cells appear in both predicted fluorescent channels, leading to ambiguous viability classification. Since viability is ultimately defined as the ratio of live to total cells, cell counting must still be performed on the predicted fluorescence images, meaning an additional postprocessing step is required. An alternative strategy formulates viability assessment as a cell detection and classification problem, where individual cells are first localized and subsequently classified. This paradigm has been applied to different imaging modalities, including second harmonic generation and autofluorescence microscopy~\cite{Chen:21}, optical coherence microscopy~\cite{Park:22}, brightfield imaging~\cite{Eren:22, Li:23}, phase imaging~\cite{Serafini:25}, 2D light scattering~\cite{Li:21}, differential interference contrast~\cite{Centofanti:25}, and digital holographic imaging~\cite{ Verduijn:21}. Despite recent progress, existing studies are mostly evaluated on limited datasets covering a narrow range of viability conditions~\cite{Chen:21, Park:22, Eren:22, Li:23, Li:21, Verduijn:21}, are often trained and validated on stained samples~\cite{Eren:22, Li:23}, and frequently rely on complicated optical setups. 

In addition, the assumption that unstained images are sufficiently similar to stained ones has rarely been examined. This is particularly important as ground-truth viability labels can only be obtained from stained samples, so models must be trained on stained images and then applied to unstained ones to enable label-free assessment. If staining alters cell morphology, the model's ability to generalize from stained to unstained images will be limited.

In this work, we present ViabiLens, a deep-learning-based pipeline for predicting the viability of label-free cells directly from brightfield images. We first examine whether brightfield images of stained cells contain sufficient information to distinguish live cells from dead ones. Building on this, we develop two supervised models: a cell detection model for localizing individual cells and a CNN classifier for assigning viability labels to each detected cell, both trained on stained samples. A key underlying assumption is that staining does not fundamentally alter the brightfield appearance of cells, allowing a model trained on stained cells to generalize to unstained ones. We first validate this assumption by analyzing the image features of individual stained and unstained cells, showing that the two populations exhibit closely matching distributions. We then test our pipeline on paired stained and unstained samples withheld from training, spanning a range of viability conditions. On CHO cells, our pipeline achieves a mean absolute error of 2.33\% against fluorescence-based measurement. Repeating the experiments under different imaging modalities, e.g. phase imaging across different focal planes, gives consistent results. These findings demonstrate the generalizability of our approach and show that unstained transmitted light images can be used directly for label-free viability prediction.

\section{Methods}
\label{sec:methods}

\begin{figure}[t]
    \centering
    \includegraphics[width=0.95\linewidth]{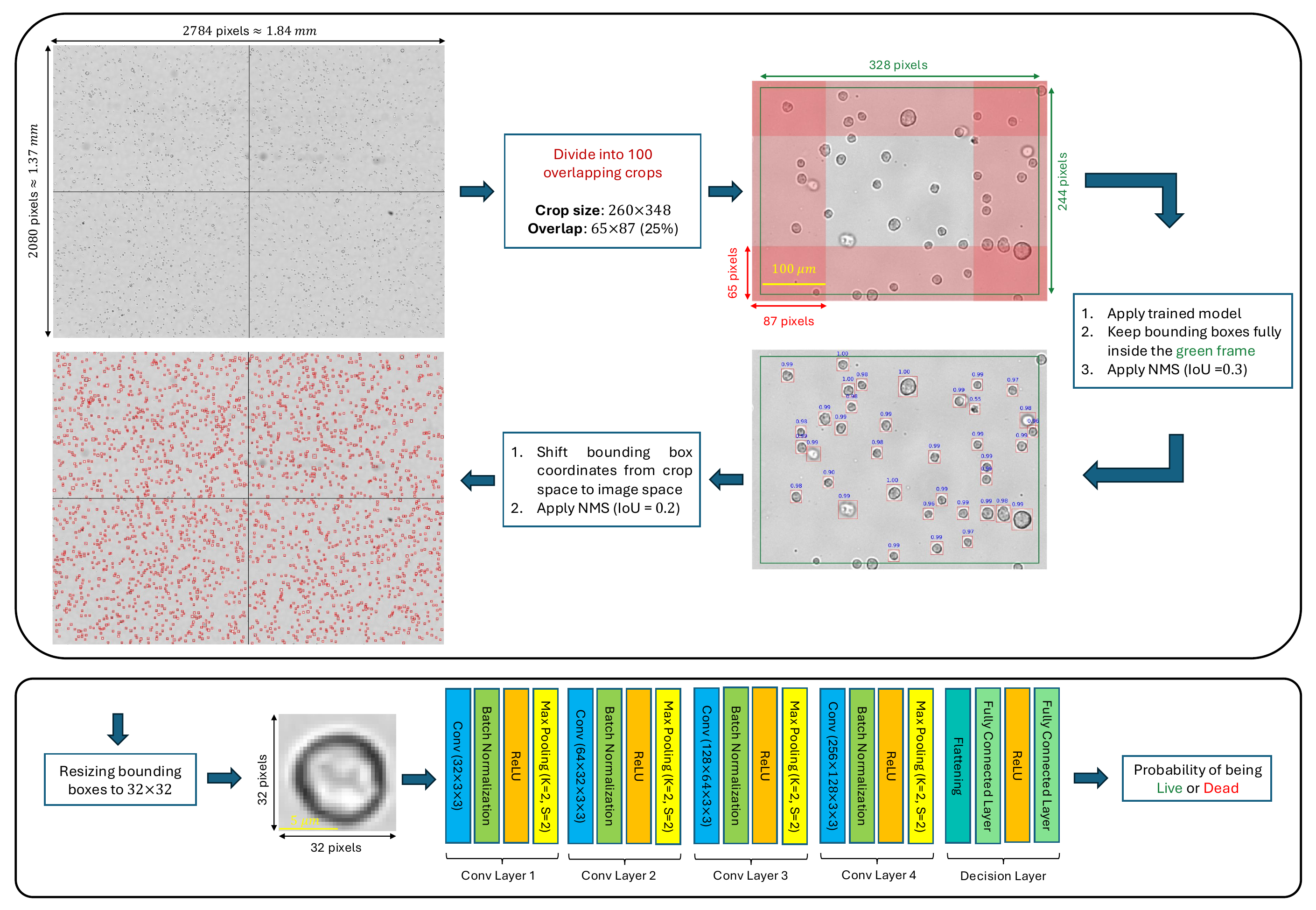}
    \caption{Overview of the cell detection and viability analysis pipeline. The brightfield image is divided into 100 overlapping crops (25\% overlap in each direction), with red-shaded regions indicating overlap with adjacent crops. Each crop is independently fed into a trained model, and only bounding boxes falling fully within the green frame are retained to discard incomplete cells near boundaries. Non-Maximum Suppression (NMS) with an Intersection over Union (IoU) threshold of 0.3 is then applied to remove redundant detections within each crop.
    Bounding box coordinates are then shifted to full image space, where a second NMS (at $IoU=0.2$) eliminates duplicate detections arising from cells appearing in red-shaded regions. Each detected cell is resized to 32$\times$32 pixels and classified as live or dead by the CNN model.}
    \label{fig:method}
\end{figure}

The overall methodology is shown in Fig.~\ref{fig:method}. Our pipeline consists of two stages: an object detection model for localizing individual cells, followed by a CNN classifier to predict their viability. We first describe the image acquisition procedure, including sample preparation, imaging settings, and the train/test split. We then detail the cell detection and classification models, followed by the evaluation metrics used to assess each stage.

\subsection{Sample Preparation}
\label{sec:sample_preparation}
Training and test data were obtained from two independent culture runs performed at different times. In both runs, NISTCHO, a clonal CHO-K1 cell line producing cNISTmAb (NIST, SKU 8675), was thawed and cultured in 30~mL of EX-CELL\textregistered{} CD CHO Fusion medium (MilliporeSigma, cat.~14365C) in a 125~mL shake flask at 37~$^\circ$C, 135~rpm (25~mm throw), 80\% relative humidity, and 5\% CO$_2$. Cells were passaged every 3--4 days by dilution into fresh medium and seeded at 0.3--0.4~$\times 10^{6}$ cells/mL. For the training set (first run), to obtain samples spanning the range of viabilities typically encountered in bioprocessing, fed-batch cultures were inoculated at 0.8~$\times 10^{6}$ cells/mL in EX-CELL\textregistered{} Advanced CHO Fed-batch Medium (MilliporeSigma, cat.~14365C) or Fusion medium in 2~L shake flasks with an initial working volume of 492~mL.
Starting on day~3 and every other day thereafter, EX-CELL Advanced CHO Feed~1 (MilliporeSigma, cat.~24368C) was added at 5\% (v/v), and a 400~g/L glucose solution (MilliporeSigma, cat.~G7021) was fed daily to restore the glucose concentration to 6~g/L. Cultures were maintained for 14 days, with samples taken at multiple time points for imaging. For the test set (second run), a portion of the passaging culture was removed from the shaking platform and incubated at 37~$^\circ$C overnight to induce cell death. This dead-cell sample was mixed with healthy cells from the same passaging culture at various ratios to produce populations with nominal viabilities of 70\%, 80\%, 87\%, 95\%, and 97\%. 

\subsection{Image Acquisition}
\label{sec:imaging}
Cells were mixed 1:1 (v/v) with ViaStain\texttrademark{} AO-PI staining solution (Revvity, cat.~no.~CS2-0106-5ML) and incubated for 10 minutes before imaging. Images ($2080 \times 2784$ pixels, corresponding to a $1.37 \times 1.84~\mathrm{mm}$ field of view) were acquired with the Cellaca\texttrademark{} PLX image cytometer from two separate, independent CHO cell culture runs performed at different times. The training set consisted of 45 stained images from the first run. From the second run, 15 stained images were acquired, each paired with a corresponding unstained brightfield image taken immediately before dye addition. These 15 stained/unstained pairs formed a held-out test set and were not used at any stage of training. For stained samples, the PLX instrument provides (i) one brightfield image and two fluorescence images, acquired in separate channels to detect AO and PI, respectively (470/532~nm excitation/emission at 15~ms exposure for AO; 531/655~nm at 40~ms for PI), as shown in Fig.~\ref{fig:overview}, (ii) cell-level bounding boxes, and (iii) live/dead viability labels for each cell. Unstained images consist of a brightfield image only, with no fluorescence channels or annotations. We use the PLX-provided bounding boxes and viability labels for training.

\subsection{Cell Detection}
\label{sec:detection}

To isolate individual cells in transmitted-light images, several approaches have been proposed. Before the deep learning era, classical image-processing methods were the primary tools for cell detection and segmentation, relying on fixed pipelines such as edge detection and morphological processing (e.g. the Watershed approach~\cite{Vincent:91}). However, these algorithms often struggle in the presence of overlapping cells. More recently, deep learning-based methods have emerged as a more robust alternative. In particular, instance segmentation and object detection models can directly predict pixel-level masks or bounding boxes for individual cells. Architectures such as Mask R-CNN~\cite{MaskRCNN:17}, Faster R-CNN~\cite{FasterRCNN:15}, YOLO~\cite{yolo:16}, and DETR~\cite{DETR:20} are prime examples of this approach, enabling more accurate separation of overlapping cells.

Because our PLX dataset provides bounding boxes, we trained an object detector (using the RF-DETR architecture~\cite{RF-DETR:26}) rather than an instance segmentation model. We used RF-DETR-Large, which has a DINOv2~\cite{DINOv2:24} backbone, initialized with weights pretrained on the COCO dataset~\cite{coco:15} and fine-tuned for binary detection (cell vs.\ background). Training directly on full-resolution ($2080 \times 2784$) images led to poor performance due to the large image area and relatively small cell size. We therefore divided each image into 100 overlapping crops of size $260 \times 348$ pixels. The training set consisted of 45 stained images, resulting in a total of 4,500 crops. Bounding-box annotations provided by PLX were used as ground-truth labels. Crops were resized to $704 \times 704$ pixels before being passed to the model. The model was trained for 20 epochs with a batch size of 4 and a learning rate of $1 \times 10^{-4}$ with AdamW optimizer~\cite{AdamW:19}. For comparison, we also trained two other detectors on the same crops: Faster R-CNN~\cite{FasterRCNN:15} with a ResNet-50-FPN~\cite{ResNet:15} backbone and the small variant of YOLO26~\cite{yolo26:26}, both initialized with COCO-pretrained weights~\cite{coco:15}. Full training details for all three architectures are provided in Supplement 1.

The inference process is shown in Fig.~\ref{fig:method}. At inference, each image was divided into 100 overlapping crops matching the training resolution, with 25\% overlap in both dimensions to ensure that cells near the boundaries appeared fully within at least one crop. Each crop was independently fed into a trained model, and only bounding boxes falling fully within the central safe zone were retained to discard incomplete cells near the crop boundaries. Since object detectors typically return multiple overlapping bounding boxes per object, NMS~\cite{Neubeck:06} (with $IoU=0.3$) was applied to remove redundant detections within each crop. Bounding box coordinates were then shifted to the full image space. 

Since a cell may appear in the overlapping regions of adjacent crops, a second NMS (with $IoU=0.2$) was applied to eliminate duplicate detections between crops. To maintain fixed-size inputs for the classification models, each detected cell was resized to $32\times32$ px and fed into the CNN. This size was selected because 95\% of cells were $19\times18$ px or smaller, while preserving aspect ratio and minimizing distortion for most cells. Anti-aliasing was applied during resizing to avoid downsampling artifacts for larger cells.

\subsection{Cell Classification}
\label{sec:classification}
We trained a compact CNN classifier (Fig.~\ref{fig:method}) on 112,947 annotated cells (101,365 live and 11,582 dead) from 45 stained images to predict cell viability. The dataset was split into 80\% training and 20\% validation, preserving the label distribution across both subsets. It was trained using cross-entropy loss and the Adam optimizer~\cite{Adam:14} with a learning rate of $10^{-6}$ for 30 epochs. The CNN has $651,234$ learnable parameters and requires 14.96 million FLOPs, with an inference time of approximately $0.50$ ms per sample on an NVIDIA A100 GPU. Once the model is trained, it can be applied to individual cell patches from either stained or unstained images. Since the model outputs a two-element softmax vector (probability of live vs. dead), the final label is assigned with a 0.5 decision threshold.

\subsection{Evaluation Metrics}
\label{sec:metrics}

We provide separate evaluation metrics for both cell detection and classification stages of the pipeline, along with an end-to-end evaluation metric for viability analysis on unstained samples. For cell detection on stained samples, we report precision, recall, F1-score, and mean Average Precision (mAP) at an IoU threshold of 0.5 using PLX bounding box annotations as the reference (exact definitions in Supplement 1). We also compare methods based on total cell counts. Since reference bounding boxes are not available for unstained images, evaluation on these samples is limited to total cell counts and qualitative visual comparison.

Classifier performance is evaluated on both the training and validation sets using the confusion matrix, overall accuracy, precision, recall, and F1-score. Due to the class imbalance ($89\%$ live cells), precision, recall, and F1-score are computed with respect to the dead (minority) class. Additionally, we report the receiver operating characteristic (ROC) curve, area under the curve (AUC), and the precision--recall curve.

Sample-level viability is quantified as the number of live cells divided by the total number of cells.  We applied our full pipeline to 15 stained and unstained image pairs. Since PLX provides viability measurements only for stained samples, we treat PLX measured viability of each stained image as the  reference for its corresponding unstained pair. Overall performance is summarized by the mean absolute error (MAE) and coefficient of determination ($R^2$ score).


\section{Results}
We first perform a calibration run using stained cells to determine whether live and dead cell populations exhibit distinct signatures under brightfield imaging. We then perform the same analysis to show that stained and unstained cells share sufficiently similar image features to allow a model trained on stained samples to generalize to unstained ones. We then present results for the cell detection, cell viability classifier, and an end-to-end evaluation of our pipeline on unstained samples. Finally, we present ViabiLens, an interactive software tool  integrating all components of the proposed pipeline for end-to-end and label-free cell viability analysis.

\label{sec:results}

\subsection{Live and Dead Cell Discrimination in Brightfield Imaging}
Our hypothesis is that live and dead cell populations exhibit sufficient differences in brightfield images that they can be distinguished by their morphology alone. To test this, we use unsupervised machine learning to compare individual cell images and sort them into representative groups. For simplicity of presentation, we map the original high-dimensional space of the images into a 2D scatter plot. After running the algorithms, interpretation of the resulting clusters is assisted by applying the PLX labels for cell detection and viability labeling. 

We applied three dimensionality reduction algorithms to the cell populations (vectorized and standardized prior to dimensionality reduction): Principal Component Analysis (PCA)~\cite{pearson:1901}, t-distributed Stochastic Neighbor Embedding (t-SNE)~\cite{tsne:08}, and Uniform Manifold Approximation and Projection (UMAP)~\cite{UMAP:20}. PCA projects data onto directions of maximum variance, providing a linear view of the feature space. t-SNE and UMAP are nonlinear methods that preserve both local and global structure by grouping similar samples and separating dissimilar ones in a low-dimensional embedding~\cite{islam:26}. UMAP has been used for similar tasks, including rare cell population detection~\cite{Weijler:21}, RNA sequencing visualization~\cite{Becht:19}, arrhythmia detection in cardiac waveforms~\cite{vazifeh:26}, and phenotyping microglial cells~\cite{Colombo:22}.

\begin{figure}[t]
    \centering
    \includegraphics[width=0.95\linewidth]{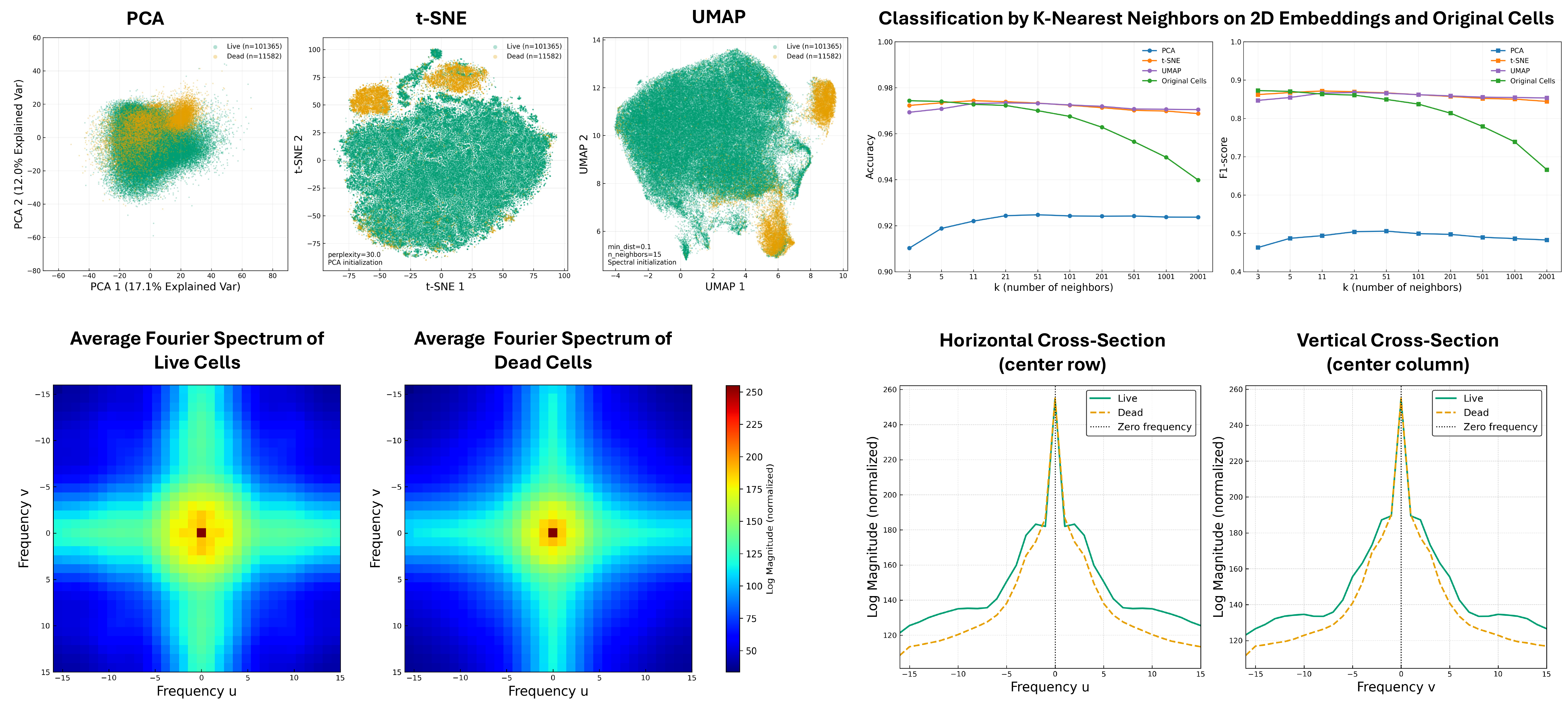}
    \caption{Comparison of live and dead cells under brightfield imaging. Top row: Two-dimensional embeddings obtained via PCA, t-SNE, and UMAP, computed from $112{,}947$ stained cells captured from $45$ training samples. Live cells are shown in green and dead cells in orange. A k-NN classifier is applied to the 2D embeddings and original cells for live vs. dead classification, as shown in the rightmost panel. Bottom row: Average Fourier spectra of live and dead cell patches (left two panels) and their horizontal and vertical cross-sections through the frequency origin (right two panels).}
    \label{fig:live-dead}
\end{figure}

Results of the 2D embeddings are shown in Fig.~\ref{fig:live-dead}. To create a 2D plot using PCA, we keep only the first two principal components (a poor cutoff in this case, as they capture only $\sim$30\% of the total variance). In the raw output, i.e. without added labels, the points overlap, and there is no way to distinguish live vs. dead cells; adding labels reveals their positions, showing that the cells below the diagonal are mostly live. t-SNE and UMAP, however, produce much more noticeable separation between live and dead cell populations, due to their ability to preserve local nonlinear structure that a linear mapping cannot capture.  

A common way to quantitatively assess the separation is to train a $k$-nearest-neighbors (k-NN) classifier directly on the 2D embeddings, which assigns each point the majority label among its $k$ nearest neighbors in the embedding space~\cite{UMAP:20}. This measures how well live and dead cells are grouped within their own population and separated from the other in the 2D space. t-SNE and UMAP achieve an accuracy of up to $\sim$97\% and an F1-score of $\sim$0.85 across a range of $k$ values. We also applied the same k-NN classification to the original high-dimensional cell representations, which yielded comparable performance to t-SNE and UMAP, indicating nonlinear embeddings largely preserve the local structure relevant to distinguishing live and dead cells. Together, these results indicate that live and dead cells occupy distinct regions of the image-feature space, and this low-dimensional separation reflects the structure of the original high-dimensional cell representations.

To further characterize the image features of live and dead populations, we computed the average Fourier spectrum across all live and dead cells~\cite{Park:22}. As shown in Fig.~\ref{fig:live-dead}, live cells exhibit markedly higher frequency content than dead cells, particularly along the principal frequency axes. This is consistent with the biological expectation that live cells possess sharper, more well-defined membrane boundaries and denser intracellular structure, both of which contribute to stronger high-frequency components in the Fourier domain. Dead cells exhibit smoother and more diffuse intensity profiles, especially in the tails of the distribution, reflecting membrane degradation and loss of structural integrity.

\begin{figure}[t]
    \centering
    \includegraphics[width=0.95\linewidth]{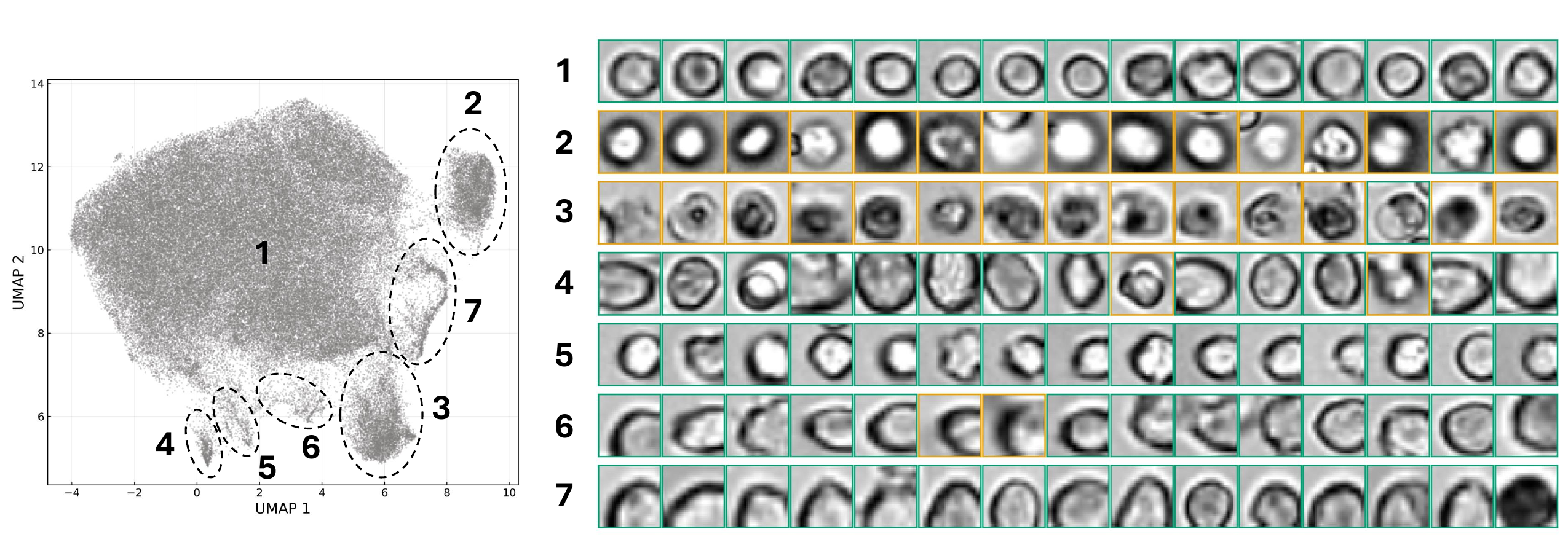}
    \caption{Unsupervised phenotyping of live and dead cells via UMAP. Left: UMAP embedding circled by cluster, with clusters manually determined based on visual separation. Right: representative brightfield cell patches from each cluster. Green and orange frames around each cell represent live and dead, respectively.}

\label{fig:Live-Dead-Phenotyping}
\end{figure}

Closer inspection of the UMAP embedding in Fig.~\ref{fig:Live-Dead-Phenotyping} reveals that the dead-cell population decomposes into two morphologically distinct sub-clusters. The first, labeled cluster~2 in Fig.~\ref{fig:Live-Dead-Phenotyping}, consists of cells that appear predominantly white and featureless, with the membrane outline still visible but a marked loss or degradation of intracellular content. The second, cluster~3, comprises cells with a heterogeneous appearance and a visible but less distinct membrane outline. We speculate that cluster~2 may reflect a later stage of cell death, in which membrane permeabilization has led to substantial loss or degradation of intracellular content, producing a homogeneous appearance. Cluster~3, which retains a faint but discernible membrane outline, is broadly consistent with reports that membrane-compromised cells appear as a blurred halo in brightfield imagery~\cite{Gisela:21}, and may represent an earlier or otherwise distinct stage of membrane degradation. 

The ability to distinguish morphologically distinct dead-cell subpopulations highlights a key advantage of unsupervised representation learning: unlike supervised classifiers, which collapse all dead cells into a single label, unsupervised embeddings preserve morphological diversity, enabling cell phenotyping without labels. Live cells (cluster~1), by contrast, have sharp membrane boundaries and moderate intracellular texture. The UMAP embedding also exposes several clusters of irregular crops (clusters~4--7). These predominantly contain partially cropped cells, in which a portion of the cell body falls outside the patch boundary. Such truncation may arise from detection errors, where the predicted bounding box fails to fully enclose the cell, or from overlapping cells, where the bounding box  captures part of an adjacent one. The separation of these malformed samples from the main clusters demonstrates that UMAP can effectively identify outlier samples in the dataset~\cite{outlier:24, vazifeh:26}. Removing these samples improves dataset quality and leads to more consistent training.

\subsection{Similarity of Stained and Unstained Cells in Brightfield Imaging}

The unsupervised approach allows us to investigate a key assumption underlying all previous imaging in cell classification:  that unstained brightfield cell images sufficiently resemble their stained counterparts, such that a model trained on stained images can generalize to unstained ones. (Proving this also motivated our use of fluorescent staining to generate cell viability ground truth, rather than visible exclusion stains such as trypan blue or erythrosin B, which would dramatically alter the brightfield appearance of the cells.) For confirmation, we collected equally sized populations of unpaired stained and unstained individual cells, where cell extraction in both conditions was performed using our trained model to ensure consistency. Although the stained and unstained images come from the same bioreactor sample, no one-to-one correspondence between individual cells across the two conditions can be established. This is because adding the stain and mixing disturbs the cells' positions within the well, making it impossible to track and image the same individual cells before and after staining. 

\begin{figure}[t]
    \centering
    \includegraphics[width=0.95\linewidth]{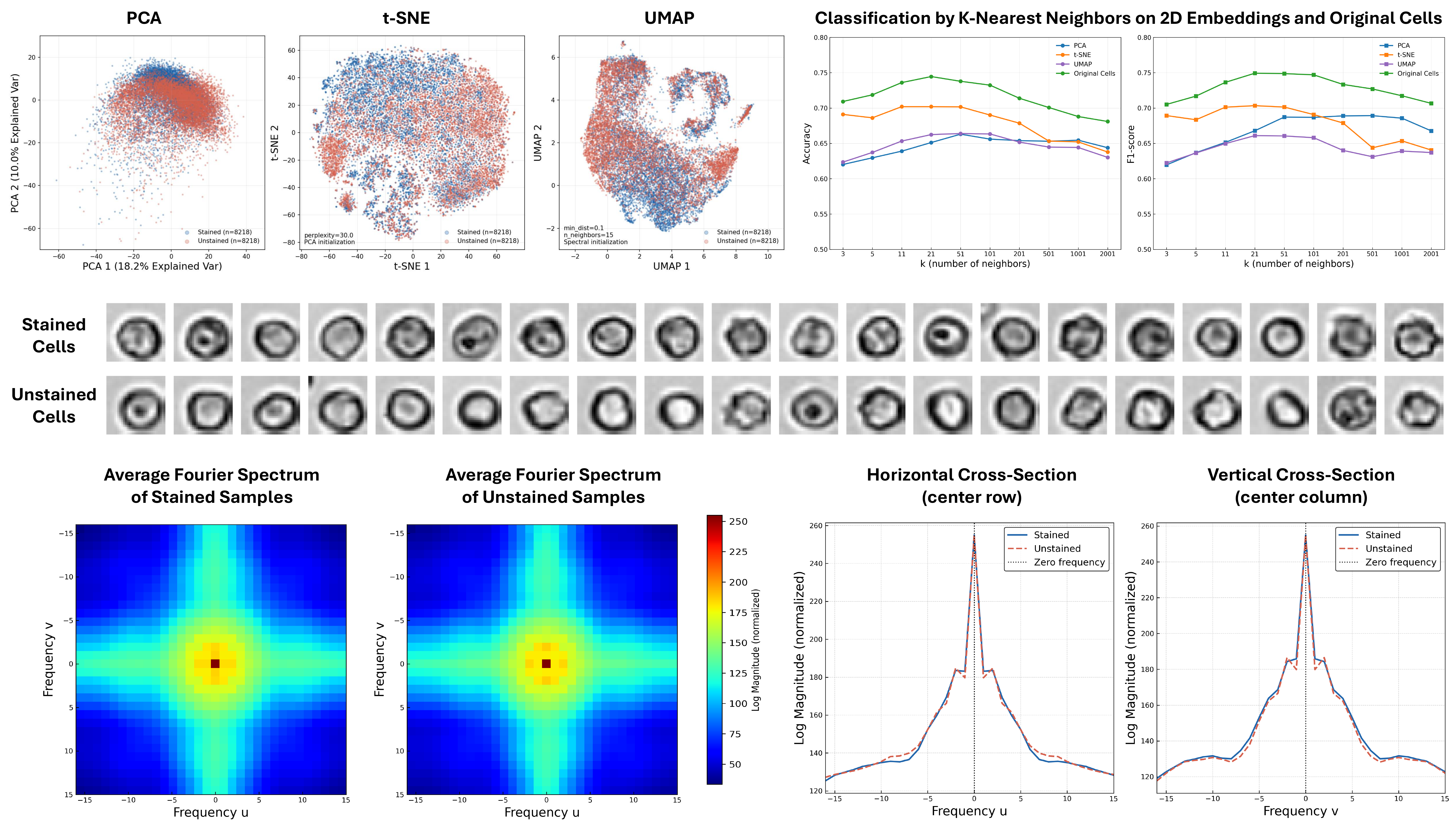}
    \caption{Comparison of stained and unstained cells under brightfield imaging. Top row: 2D embeddings obtained via PCA, t-SNE, and UMAP, computed from 16,436 cells captured from 15 stained and unstained image pairs. Unstained patches were randomly subsampled to match the stained count; the two sets are not paired one-to-one. Middle row: 20 examples of stained and unstained cells. Bottom row: Average Fourier spectra of stained and unstained cells (left two panels) and their horizontal and vertical cross-sections through the frequency origin (right two panels).}
    \label{fig:stained_unstained}
\end{figure}

As shown in Fig.~\ref{fig:stained_unstained}, applying PCA, t-SNE, and UMAP to the cell populations produces intermixing of stained and unstained cells in the projected 2D space. This indicates that the two populations occupy overlapping regions of the feature space. To quantify this, we trained the same k-NN classifier used in our live/dead analysis to distinguish stained from unstained cells. Across all four representations, including the original high-dimensional cells, classification accuracy remains low ($\sim$60--70\%), well below the $\sim$97\% accuracy achieved on the live/dead task in Fig.~\ref{fig:live-dead}. Likewise, the average Fourier profiles of stained vs. unstained cells are nearly identical. Taken together, the results show that AO and PI staining does not substantially alter cell appearance under brightfield imaging. 

\subsection{Cell Detection Performance}
\begin{figure}[t]
    \centering
    \includegraphics[width=0.95\linewidth]{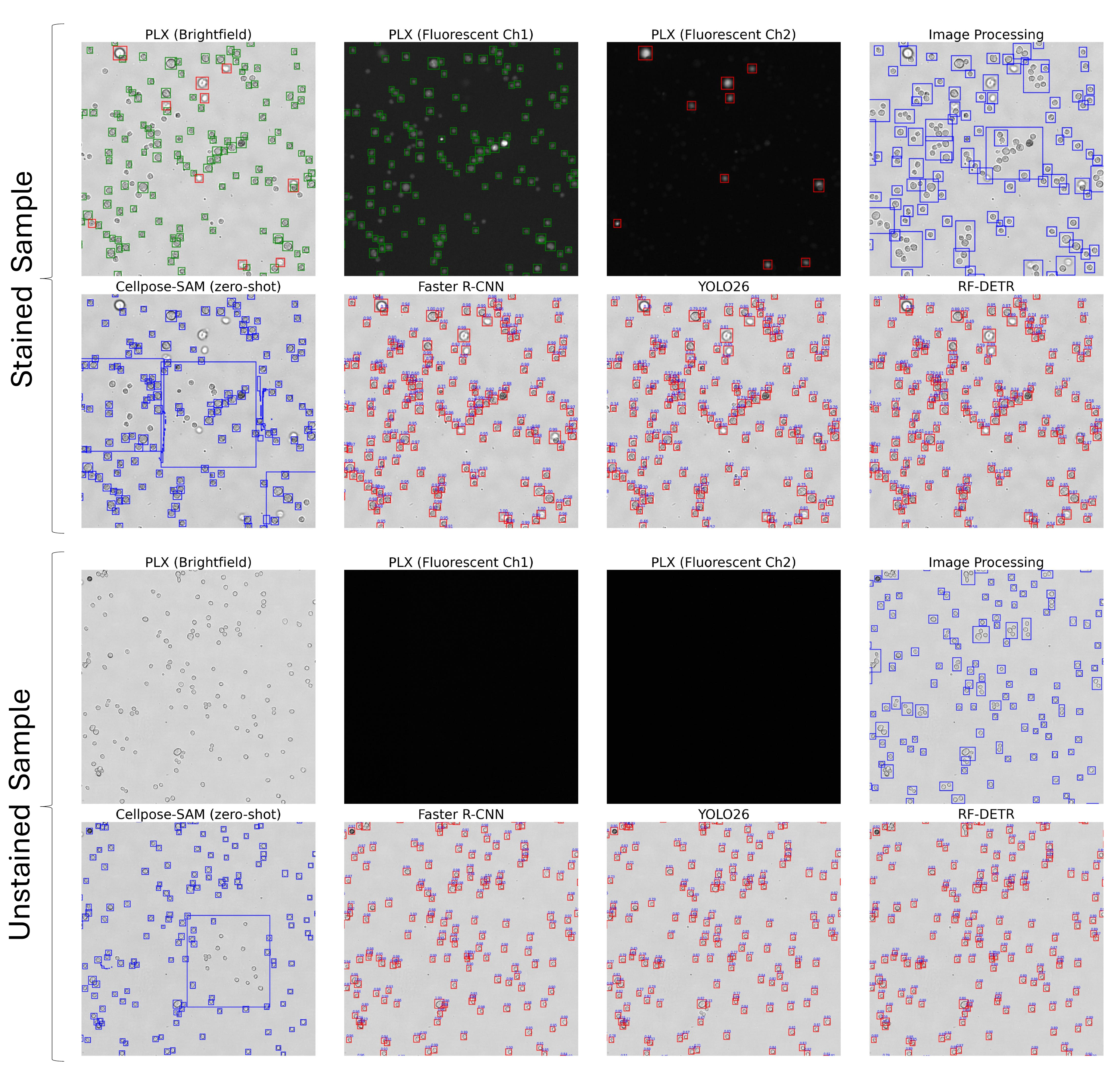}
    \caption{Qualitative comparison of cell detection methods on stained (top) and unstained (bottom) samples. For each sample, the first row shows the PLX brightfield image with its reference annotations, the two fluorescent channels from which those annotations are derived (Ch1: live cells, green boxes; Ch2: dead cells, red boxes), and detections from the image-processing baseline. The second row shows detections from Cellpose-SAM (zero-shot) and the three trained detectors (Faster R-CNN, YOLO26, and RF-DETR). Each trained-model detection is labeled with its confidence score, the predicted probability that the proposal is a cell. PLX relies on fluorescence and therefore provides no annotations for the unstained sample, whose fluorescent channels are blank.}
    \label{fig:segmentation_comparison}
\end{figure}

To assess whether task-specific training is necessary, we compared our approach against baselines representing the main paradigms available for our task: (1) the Cellaca\texttrademark{} PLX image cytometer, whose detections also served as the reference annotations for training; (2) a classical image-processing pipeline that requires no training data (details in Supplement~1); and (3) the pre-trained Cellpose-SAM~\cite{Stringer:21, Cellpose-SAM:25}, a generalist foundation model applied off the shelf without fine-tuning. Our approach represents the fourth paradigm, in which detection models are fine-tuned on task-specific data; we evaluated three architectures (Faster R-CNN~\cite{FasterRCNN:15}, YOLO26~\cite{yolo26:26}, and RF-DETR~\cite{RF-DETR:26}). Fig.~\ref{fig:segmentation_comparison} shows qualitative detection results on stained and unstained images (more examples in Supplement~1). PLX provides bounding-box annotations only for stained samples --- and thus is not applicable in label-free settings --- and its system misses a noticeable number of cells. However, these cells still exhibit fluorescence signals, suggesting that the omissions are likely due to limitations of the detection algorithm rather than any biological difference. The image-processing pipeline performs well on isolated cells but struggles to accurately resolve overlapping or aggregated cells (known as cell clumping). On the other hand, localizing and studying cell clumps may itself be of interest, and image processing can provide bounding boxes for these scenarios. Despite its limitations at the instance level, it can also provide a lower bound on the total number of cells in the sample (as all cells within a clump are counted as a single detection). Cellpose-SAM~\cite{Cellpose-SAM:25} struggles with cells exhibiting a white appearance, which are primarily dead cells, and often produces inaccurate segmentation, including overly large merged regions or excessively small spurious detections. In contrast, our trained models can detect cells even within tightly clustered regions. Further, they provide a confidence score for each prediction, which can be used to filter out low-confidence detections and improve robustness. Visually, we observed no noticeable difference among the three architectures.

Table~\ref{tab:cell_detection_comparison} reports detection performance for all methods on stained and unstained images. For stained samples, we report cell counts, precision, recall, F1-score, and mAP; for unstained samples we report only cell counts due to the absence of reference annotations. For cell counts, our method detects approximately 15--20\% more cells compared to the PLX measurement (on the training set). Visual inspection across different samples suggests that our method is able to recover missed or inaccurate bounding boxes. This also suggests that training on a large set of imperfect annotations may allow a deep learning model to generalize beyond individual annotation errors and recover missed cells. Cellpose-SAM detects slightly more cells than our trained models, but visual inspection indicates that some of these detections correspond to very small false-positive cells or large bounding boxes that do not correspond to individual cells. Overall, the trained models achieve the best performance across all evaluation metrics, with $mAP \geq 0.9$ on both the training and test sets.

Cell appearance can vary across focal planes, and the detection model should remain robust to such focus-dependent changes. To verify this, we evaluated the trained RF-DETR model on a focal stack acquired at a fixed field of view, with eleven different focal planes (Fig.~\ref{fig:focal_stack_main}; additional example in Supplement~1). Due to slight sample movement during acquisition, cells are not perfectly registered across planes, with some entering or leaving the field of view. Despite this, the model correctly identified all cells within each focal plane, detecting a consistent $50 \pm 2$ cells across the entire focal range, with the small variation attributable to cells entering or leaving the field of view.  The other two architectures achieved similar results.

\begin{table*}[t]
\centering
\caption{Cell detection results on stained and unstained samples. PLX serves as the reference measurement; since it only provides bounding-box annotations for stained samples, some unstained metrics are not provided.}
\label{tab:detection_results}
\label{tab:cell_detection_comparison}
\scriptsize
\setlength{\tabcolsep}{4pt}
\renewcommand{\arraystretch}{1.0}
\resizebox{\linewidth}{!}{%
\begin{tabular}{cccccccc}
\toprule
Sample & Metric & \begin{tabular}[c]{@{}c@{}}PLX\\(Reference)\end{tabular}
  & \begin{tabular}[c]{@{}c@{}}Image\\Processing\end{tabular}
  & \begin{tabular}[c]{@{}c@{}}Cellpose-SAM~\cite{Stringer:21, Cellpose-SAM:25}\\(zero-shot)\end{tabular}
  & \begin{tabular}[c]{@{}c@{}}Faster R-CNN~\cite{FasterRCNN:15}\\(fine-tuned)\end{tabular}
  & \begin{tabular}[c]{@{}c@{}}YOLO26~\cite{yolo26:26}\\(fine-tuned)\end{tabular}
  & \begin{tabular}[c]{@{}c@{}}RF-DETR~\cite{RF-DETR:26}\\(fine-tuned)\end{tabular} \\
\midrule
\multirow{5}{*}{\begin{tabular}[c]{@{}c@{}}45 stained samples\\(training set)\end{tabular}}
  & Cell count & 112,947 & 105,190 & 136,334 & 132,110 & 130,221 & 136,127 \\
  & Precision  & --      & 0.56    & 0.76    & 0.86    & \textbf{0.87} & 0.84 \\
  & Recall     & --      & 0.52    & 0.88    & \textbf{0.96} & 0.95 & \textbf{0.96} \\
  & F1         & --      & 0.54    & 0.81    & \textbf{0.90} & \textbf{0.90} & 0.89 \\
  & mAP        & --      & --      & --      & 0.92    & 0.91    & \textbf{0.93} \\
\midrule
\multirow{5}{*}{\begin{tabular}[c]{@{}c@{}}15 stained samples\\(test set)\end{tabular}}
  & Cell count & 8,056   & 7,642   & 8,304   & 8,256   & 8,247   & 8,218 \\
  & Precision  & --      & 0.81    & 0.70    & \textbf{0.91} & \textbf{0.91} & \textbf{0.91} \\
  & Recall     & --      & 0.78    & 0.72    & 0.93    & \textbf{0.94} & 0.93 \\
  & F1         & --      & 0.79    & 0.71    & \textbf{0.92} & \textbf{0.92} & \textbf{0.92} \\
  & mAP        & --      & --      & --      & \textbf{0.92} & \textbf{0.92} & 0.91 \\
\midrule
\begin{tabular}[c]{@{}c@{}}15 unstained samples\\(test set)\end{tabular}
  & Cell count & --      & 14,305  & 17,667  & 16,597  & 16,656  & 16,693 \\
\bottomrule
\end{tabular}}
\end{table*}

\begin{figure}[t]
    \centering
    \includegraphics[width=0.95\linewidth]{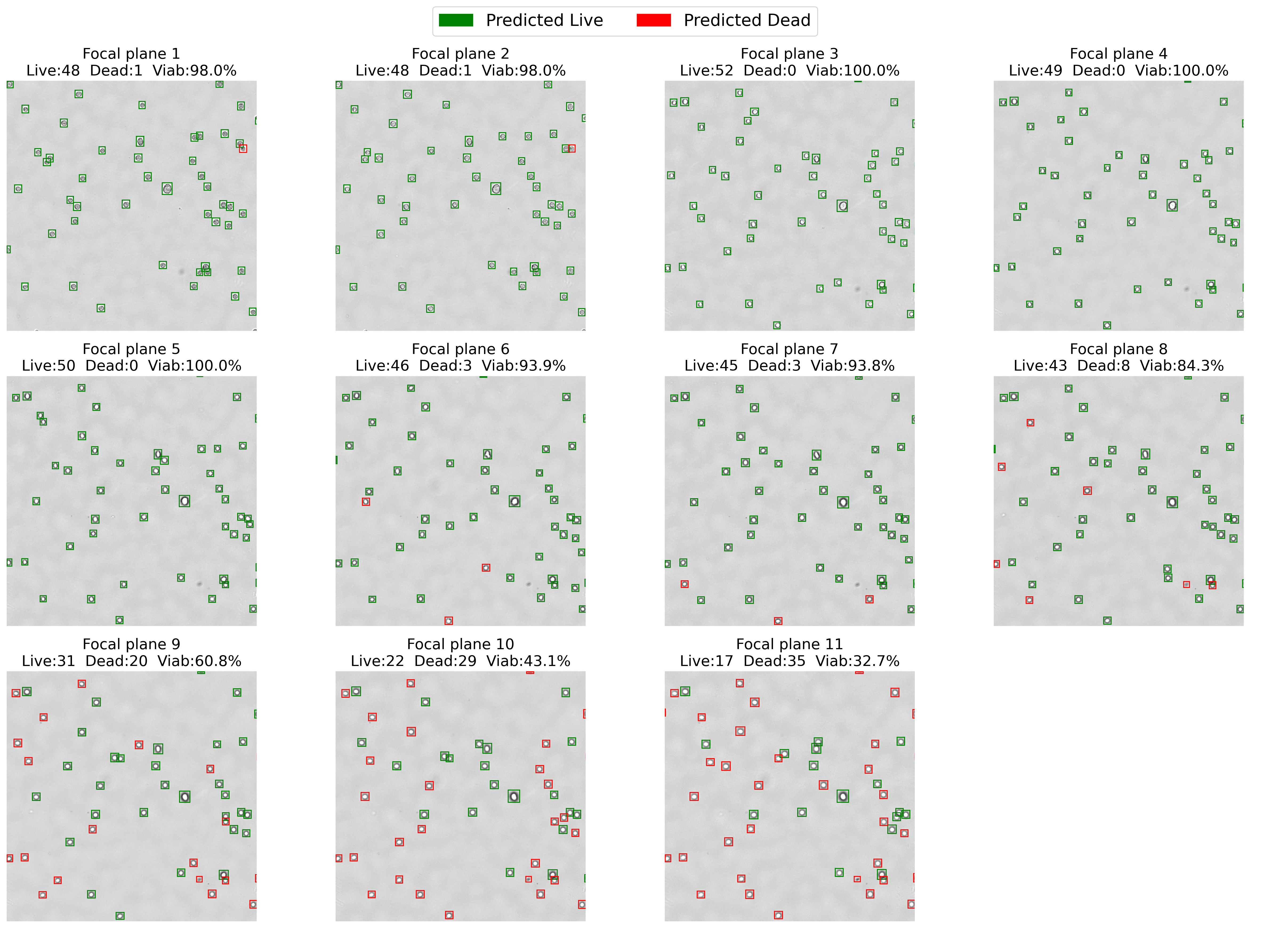}
   \caption{Performance of our pipeline on cells under eleven different focal planes. Cell appearance changes across focal planes, and while detection remains robust, the viability classifier can be more sensitive to focal planes outside its training distribution. Cells are also moving, so frames are not perfectly registrable. A video of the focal stack is available in the \href{https://amirrezavazifeh.github.io/ViabiLens-Project-Page/}{project page}.}
   \label{fig:focal_stack_main}
\end{figure}

\subsection{Cell Classification Performance}

\begin{figure}[t]
    \centering
    \includegraphics[width=0.95\linewidth]{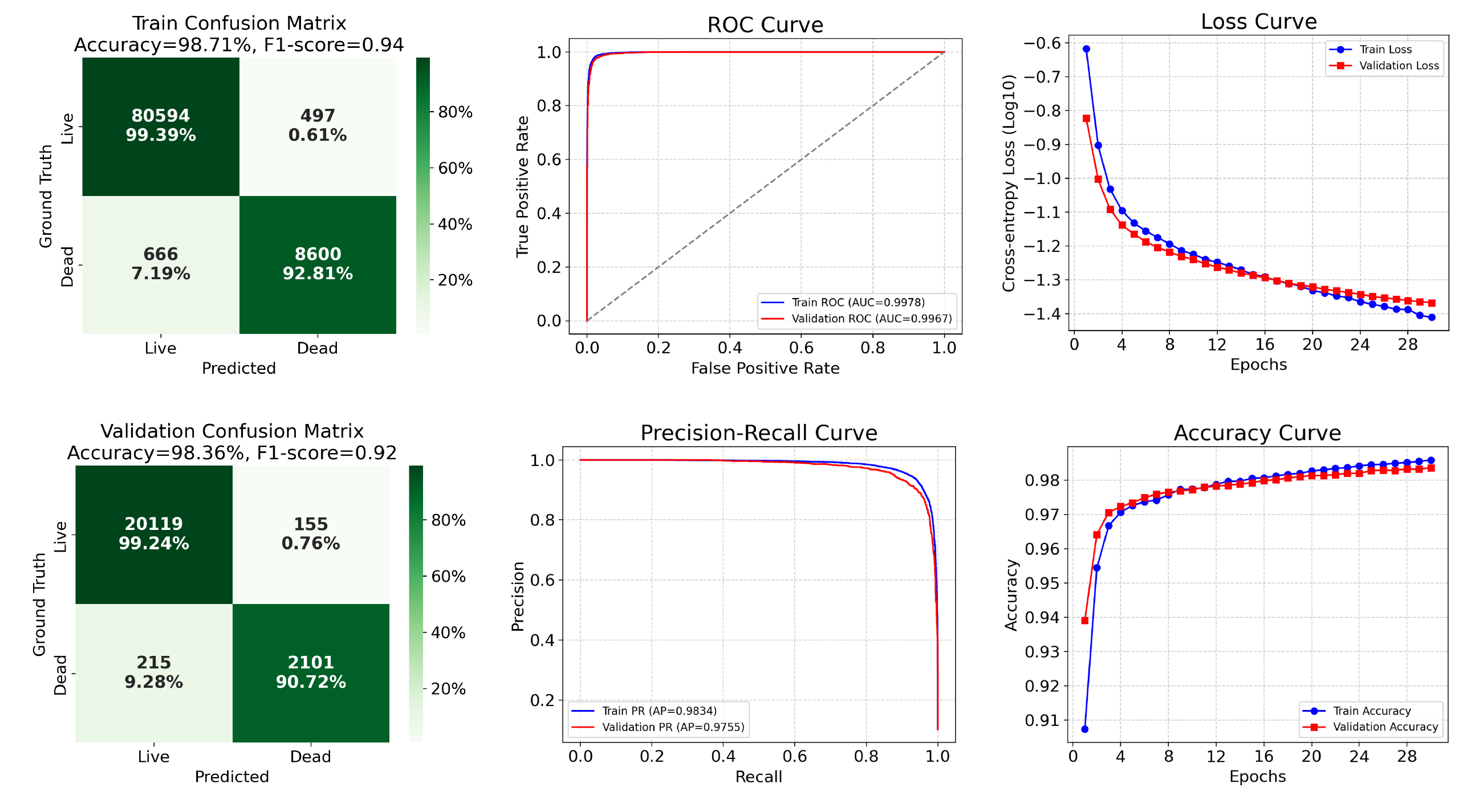}
    \caption{Performance of the binary cell viability classifier. Left: training and validation confusion matrices. Middle: ROC curve and precision-recall curve. Right: loss and accuracy curves across 30 epochs.}
    \label{fig:classification_results}
\end{figure}

As shown in Fig.~\ref{fig:classification_results}, the CNN model achieves 98.71\% accuracy (F1-score = 0.94) on the training set and 98.36\% accuracy (F1-score = 0.92) on the validation set, with AUC values of 0.9978 and 0.9967, respectively. The precision-recall curves further support strong classification performance, with average precision of 0.9834 on training and 0.9755 on validation. The loss and accuracy curves in Fig.~\ref{fig:classification_results} show close agreement between training and validation across epochs, confirming that the model generalizes well without overfitting.

\subsection{End-to-End Viability Prediction}

\begin{figure}[t]
    \centering
    \includegraphics[width=0.95\linewidth]{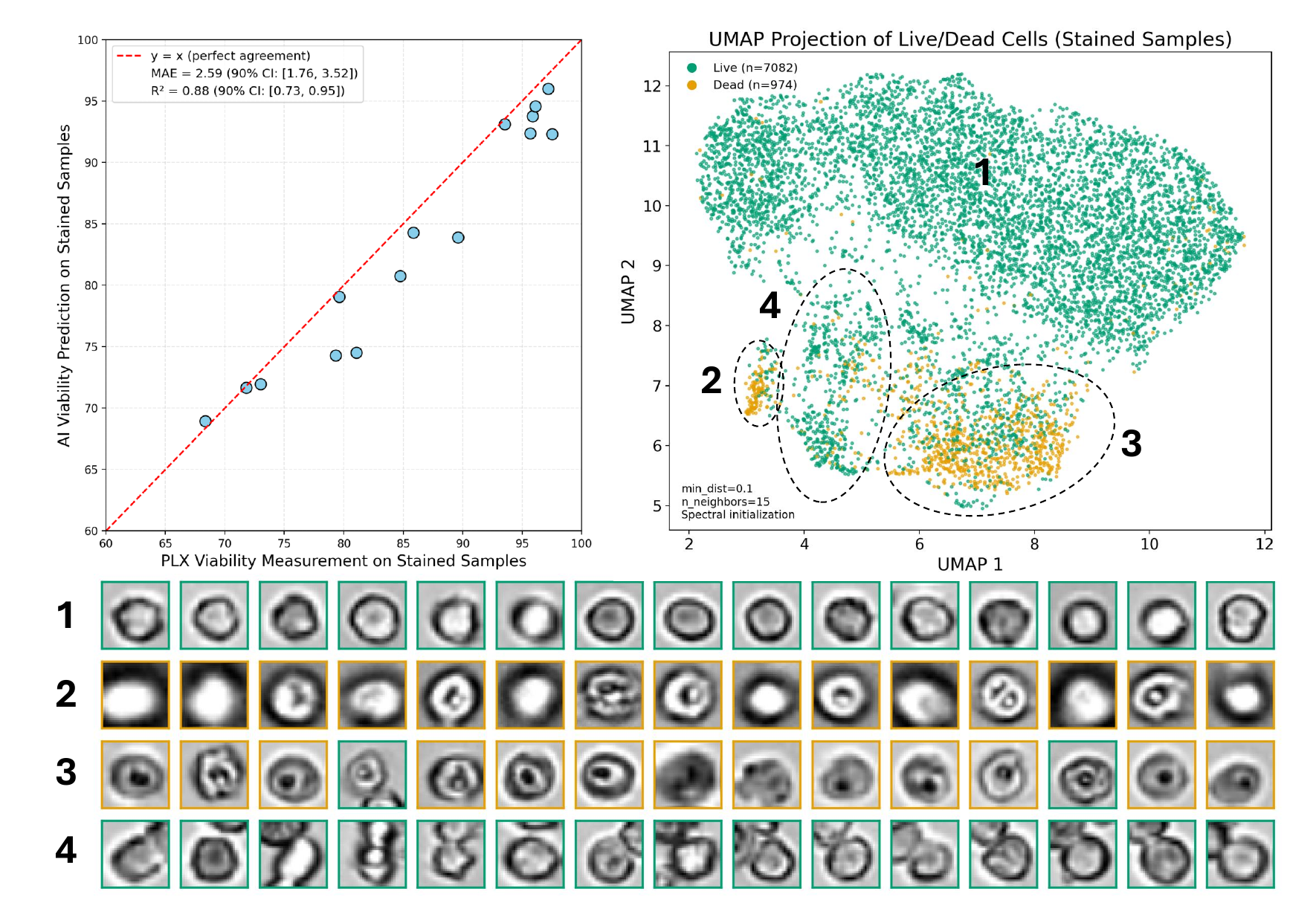}
    \caption{Sample-level viability predictions and UMAP projections for stained samples (test set). Scatter plots show predicted viability against PLX-measured viability, with the dashed line indicating perfect agreement. UMAP projections show live (green) and dead (orange) cell populations, colored by PLX-measured labels. Representative cell crops from each cluster are shown below the corresponding UMAP projection.}
    \label{fig:viability_result_stained}
\end{figure}

\begin{figure}[t]
    \centering
    \includegraphics[width=0.95\linewidth]{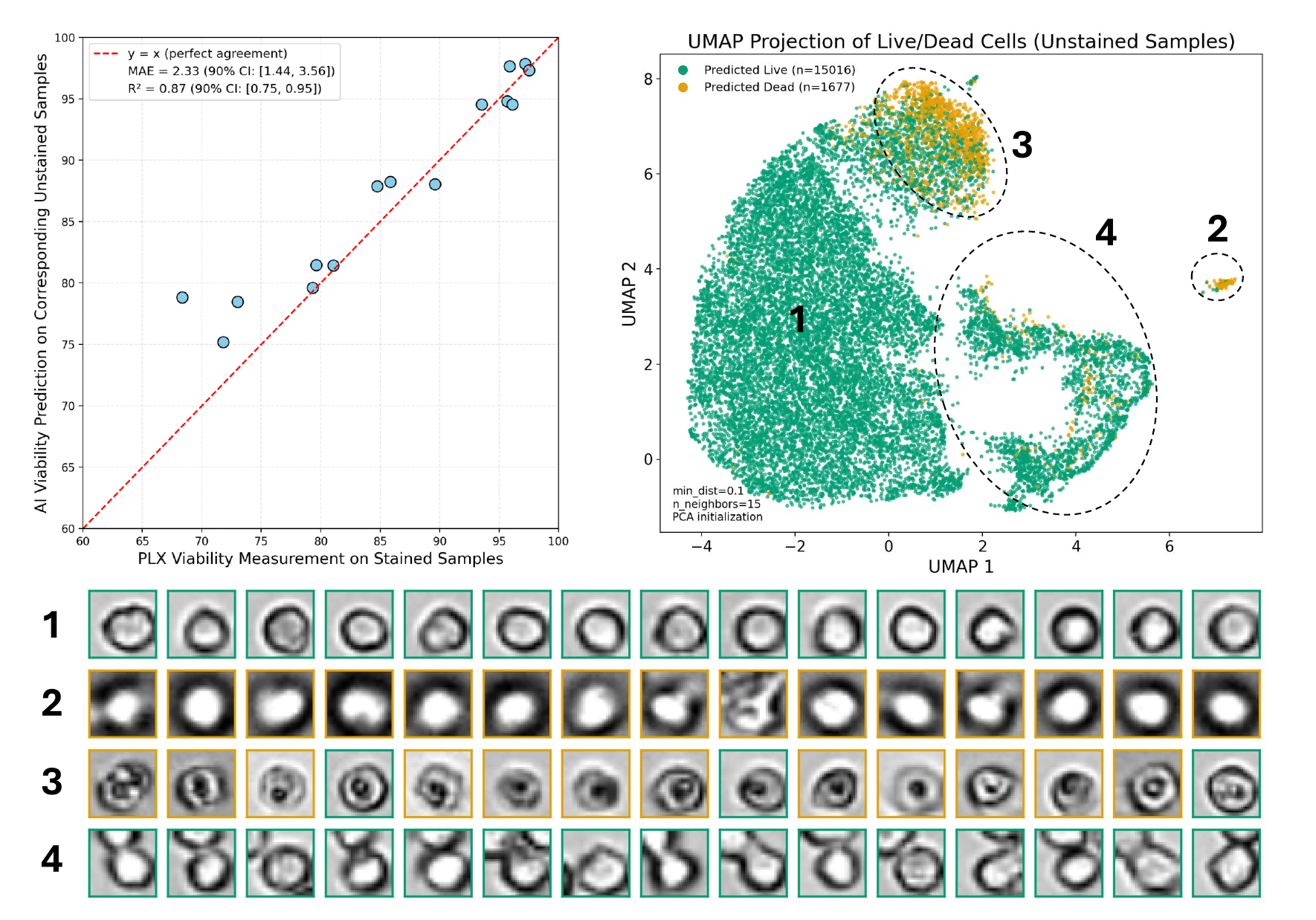}
    \caption{Sample-level viability predictions and UMAP projections for unstained samples (test set). Scatter plots show predicted viability against PLX-measured viability, with the dashed line indicating perfect agreement. UMAP projections show live (green) and dead (orange) cell populations, colored by CNN-predicted labels. Representative cell crops from each cluster are shown below the corresponding UMAP projection.}
    \label{fig:viability_result_unstained}
\end{figure}

We applied our pipeline to 15 samples with paired stained and unstained images, with PLX viability measurements on stained samples serving as reference. Sample-level viability is predicted with our pipeline, and performance is evaluated using both MAE and the $R^2$ score, with 90\% confidence intervals (CI) obtained via bootstrap resampling. Results are summarized in Figs.~\ref{fig:viability_result_stained} and~\ref{fig:viability_result_unstained}.

For stained samples, we achieved an MAE of 2.59\% (90\% CI: [1.76, 3.52]) and $R^2$ of 0.88 (90\% CI: [0.73, 0.95]), as shown in Fig.~\ref{fig:viability_result_stained}. The scatter plot shows our predictions falling slightly below PLX measurements across most samples; a potential reason is that our detection model identifies more cells than PLX, and these additional cells tend to have lower viability, which pulls down the overall sample-level estimate. We further perform a population-level analysis using UMAP. For stained samples, where reference labels are available, we apply UMAP to the cells extracted and labeled by PLX, which reveals visually separable clusters of live and dead cells. By visualizing representative examples from each cluster in the UMAP plot, we observe distinct morphological patterns: Cluster 1 consists of sharp, well-defined cells that are predominantly live; Cluster 2 contains featureless, translucent cells lacking internal structure; Cluster 3 comprises fuzzy, blurred cells that frequently contain a small dark internal punctum, which may correspond to a condensed nucleus or debris material; and Cluster 4 captures image patches containing multiple adjoining cells, likely corresponding to cells undergoing division.

For unstained samples, we achieved an MAE of 2.33\% (90\% CI: [1.44, 3.56]) and $R^2$ of 0.87 (90\% CI: [0.75, 0.95]). We conducted an analogous UMAP analysis, where cells are extracted using the trained model and labeled with the CNN classifier, as shown in Fig.~\ref{fig:viability_result_unstained}. Predicted dead cells are grouped closely together in the embedding, and again we observe four main clusters, suggesting that the CNN has learned discriminative features that generalize beyond stained samples. Examining representative unstained cells from the UMAP embedding reveals morphological patterns that closely mirror those observed in the stained samples: Cluster 1 again consists of sharp cells that are predominantly predicted live; Cluster 2 contains cells with a white appearance resembling those in stained Cluster 2 and is mostly predicted as dead; Cluster 3 comprises the majority of predicted-dead cells, which appear fuzzy and blurred; and Cluster 4 similarly captures image patches with multiple adjoining cells, likely corresponding to cells undergoing division. This correspondence suggests that the visual features distinguishing live and dead cells are largely preserved between stained and unstained imaging, supporting the CNN's ability to generalize across modalities.

\subsection{Software Development}

\begin{figure}[t]
    \centering
    \includegraphics[width=\linewidth]{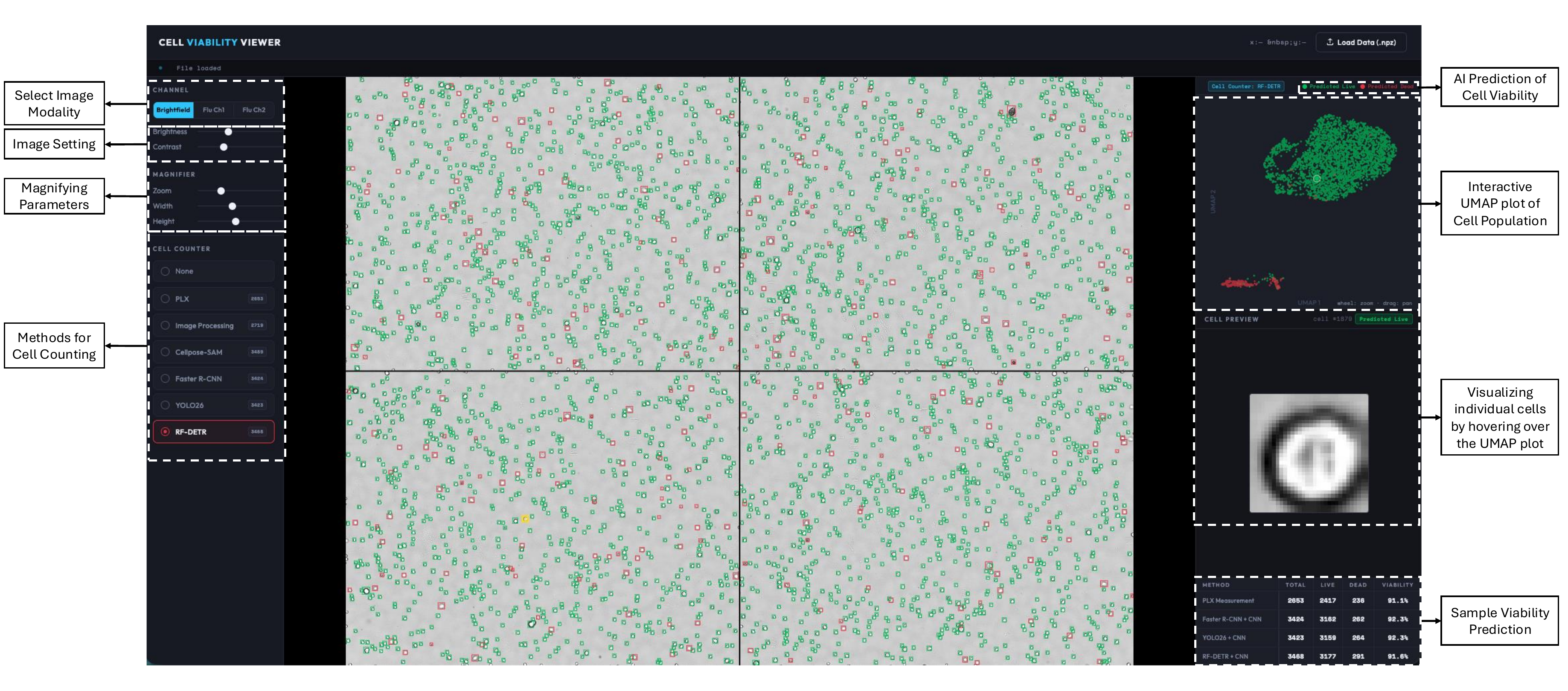}
    \caption{ViabiLens, an interactive software tool for visualizing and benchmarking cell viability analysis methods. The left panel provides controls for image modality selection, brightness, contrast, and cell counter selection. The central panel displays brightfield images with overlaid bounding boxes from each method. The right panel shows an AI-predicted UMAP embedding with a single-cell preview window and aggregate viability statistics. The demo of the software is available in the \href{https://amirrezavazifeh.github.io/ViabiLens-Project-Page/}{project page}.}    
    \label{fig:software}
\end{figure}

To facilitate assessment and interactive benchmarking of the proposed pipeline, we developed ViabiLens, an interactive HTML-based platform. As shown in Fig.~\ref{fig:software}, the interface provides side-by-side visualization of brightfield and fluorescence image modalities, with adjustable brightness, contrast, and magnification controls. Detected cells from all six methods are rendered as overlaid bounding boxes, enabling direct visual comparison of segmentation outputs. The right panel displays a UMAP projection of the analyzed sample, where AI-predicted viability labels are encoded as color-coded point clusters and hovering over individual points renders the corresponding cell crop in a preview window. Aggregate viability statistics are reported in a summary table at the bottom of the panel. Beyond its utility as a research tool for testing and refining methods, the platform is functional software with potential commercial application.

\section{Discussion}
\label{sec:discussion}
Our results demonstrate that deep-learning-based cell detection and classification can predict sample-level viability directly from unstained brightfield images, closely matching the reference measurements obtained from stained samples. Compared to virtual staining approaches, which predict fluorescence images from brightfield inputs and require cell counting and viability analysis as a subsequent post-processing step, our method produces per-cell viability predictions directly. Virtual staining also inherits an ambiguity present in physical staining: some cells appear in both fluorescent channels simultaneously, and a virtual staining model can replicate this effect, making viability classification more ill-defined. The detection-and-classification paradigm avoids these issues entirely. Furthermore, the output of the classifier provides a continuous probability of being live or dead rather than a hard binary label, offering a more quantitative and nuanced characterization of cell viability. The temperature parameter of the softmax can also be adjusted before training to control the sharpness of these probabilities, allowing more or less deterministic predictions. A similar score is available from the detection stage, where the objectness confidence can be used to filter out low-quality bounding box proposals. A natural future direction is to train on paired brightfield and fluorescence crops of single cells (e.g., virtual staining at the single-cell level) to provide richer information about cell health while remaining compatible with the detection-then-classification paradigm introduced here.

Several factors can influence pipeline performance. Focal plane is one such factor: as shown in Fig.~\ref{fig:focal_stack_main}, predictions on the same cell captured at different depths can vary, particularly for the viability classifier, which may not have been exposed to all focal planes during training. Nevertheless, a focal stack of individual cells opens the opportunity for three-dimensional reconstruction of cell structure~\cite{Pegard:13}, which could potentially improve viability classification. Bounding box accuracy is another source of variability: the same cell classified under slightly shifted bounding boxes can receive different viability labels, as the crop content changes. One way to address this is to jointly train the detector and classifier by optimizing the detector for live, dead, and background classes rather than cell versus background alone. However, since object detectors return multiple overlapping proposals per cell with each potentially receiving a different live/dead prediction, this formulation complicates downstream NMS and requires careful design. In addition, because individual cells are small relative to the original image, the trained model must be applied to smaller regions. Rather than manually cropping these regions, the bounding boxes generated by image processing can provide candidate cell regions with fewer overlapping cells, which can then be processed by the trained model.

In Supplement 1, we repeat all experiments using a higher-resolution microscope across different imaging modalities and focal planes, and results remain consistent, supporting the generalizability of our claims. When multiple modalities or focal planes are available at training time, multi-modal learning can further improve performance over any single modality. Finally, we have shown that nonlinear dimensionality reduction via UMAP enables unsupervised phenotyping of live and dead cell populations. In the multi-modal setting, manifold alignment offers a complementary tool: by mapping cell representations from different imaging conditions onto a shared low-dimensional space, it can enhance the discriminability of live and dead populations across modalities and facilitate knowledge transfer between imaging conditions~\cite{Vazifeh_manifold_alignment:26}.

\section{Conclusion}
\label{sec:conclusion}
We presented a deep-learning-based pipeline for cell viability assessment directly from unstained brightfield images. We first established that live and dead cells exhibit distinct morphological signatures under stained brightfield imaging, and that stained and unstained cells share closely matched image features. These two findings jointly justify training on stained samples and deploying on unstained ones. Our pipeline combines a cell detection model for localizing individual cells with a compact CNN classifier for live/dead prediction, achieving a mean absolute error of 2.33\% on unstained CHO cells across a wide range of viability conditions. These results open new possibilities for continuous, non-invasive monitoring of cell health.


\begin{backmatter}

\section*{Funding}
The authors gratefully acknowledge financial support from Waters Corporation.

\section*{Acknowledgments}
Portions of this work were previously presented at the Optica Biophotonics Congress 2026~\cite{Vazifeh_manifold_alignment:26} and at Frontiers in Optics + Laser Science (FiO+LS) 2026.

\section*{Disclosures}
ARV, JWF: Waters Corporation (F,P); CZ, SM, RJ: Waters Corporation (E,P), 
where F denotes financial support, E employment and P patents. 
\section*{Data Availability}
The datasets and trained models are confidential information of Waters Corporation under a research collaboration agreement and are not publicly available. Data can be made available upon reasonable request, directed to Dr. Christian Zeigler (Waters Corporation).

\bmsection{Supplemental document}
See Supplement~1 for supporting content.

\end{backmatter}

\FloatBarrier
\bibliography{references}

\clearpage
\appendix
\setcounter{figure}{0}  \renewcommand{\thefigure}{S\arabic{figure}}
\setcounter{table}{0}   \renewcommand{\thetable}{S\arabic{table}}
\setcounter{equation}{0}\renewcommand{\theequation}{S\arabic{equation}}
\setcounter{algorithm}{0}\renewcommand{\thealgorithm}{S\arabic{algorithm}}

\begin{center}
{\Large\bfseries Supplement 1:\\[2pt]
Label-free cell counting and viability prediction with brightfield imaging and deep learning}
\end{center}
\suppressfloats[t]
\vspace{1em}

This supplemental document is organized as follows. Section~\ref{cell detection} provides further details on cell detection. Section~\ref{image processing} describes the image-processing pipeline used in the main paper, and Section~\ref{cellpose} reports the settings used for Cellpose-SAM~\cite{Cellpose-SAM:25, Stringer:21} predictions. Section~\ref{Trained Model} describes the training of the three detectors (Faster R-CNN~\cite{FasterRCNN:15}, YOLO26~\cite{yolo26:26}, and RF-DETR~\cite{RF-DETR:26}). Finally, Section~\ref{Additional Results} provides further qualitative results for cell detection, along with tables underlying the figures in the paper. In Section~\ref{Evaluation Details}, we provide details of the metrics used to evaluate different stages of our pipeline. Finally, in Section~\ref{ZEISS Microscope}, we repeat the same experiment using a Zeiss microscope and achieve results consistent with the PLX experiment.



\section{Cell Detection}
\label{cell detection}

\subsection{Image Processing}
\label{image processing}
\begin{figure}[t]
    \centering
    \includegraphics[width=\linewidth]{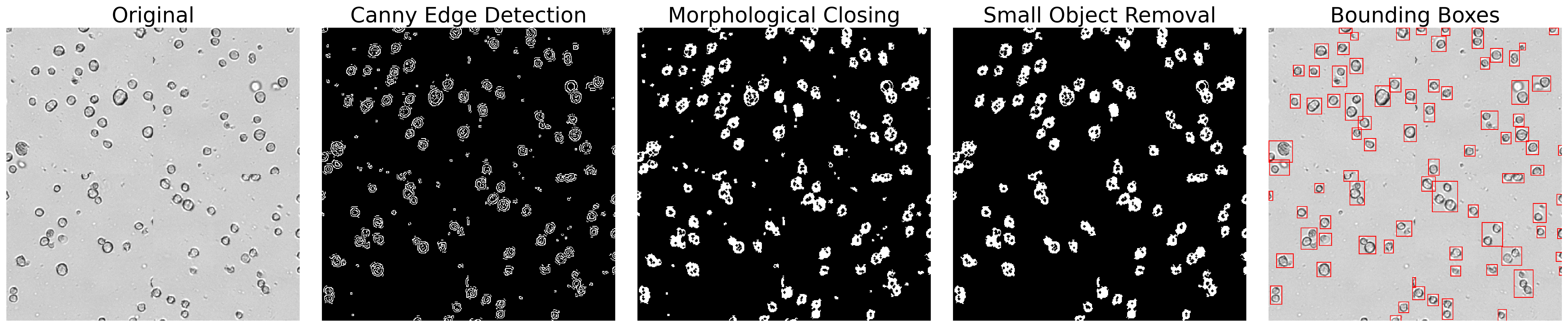}
    \caption{Stages of the image processing pipeline for cell detection on PLX data.}
    \label{fig:image_processing_plx}
\end{figure}

The different stages of the proposed image processing pipeline are shown in Figure~\ref{fig:image_processing_plx} and presented in Algorithm~\ref{alg:instance_segmentation}.  Given a brightfield image with RGB channels, we first convert it to grayscale using the first channel, as all three PLX channels contain identical information. Canny edge detection \cite{canny:86} is then applied to extract cell boundaries, followed by a morphological closing operation to fill gaps and generate a binary segmented image. To detect the boundaries of connected white regions and
delineate each candidate object, contours are subsequently identified using OpenCV's \texttt{findContours} function. Since Canny edge detection and morphological closing can also pick up small debris and high-frequency noise as spurious objects, small objects below a predefined area threshold are removed to reduce noise and debris. Bounding boxes are then drawn around the remaining objects and considered as the final cell detections.

\begin{algorithm}[t]
\caption{Cell Detection Using Image Processing}
\label{alg:instance_segmentation}
\begin{algorithmic}[1]
\Require Brightfield image $I \in \mathbb{R}^{H \times W \times 3}$
\Require Canny edge detection thresholds $\tau_{\text{low}} = 100$, $\tau_{\text{high}} = 200$
\Require Elliptical kernel size $k = 3 \times 3$
\Require Minimum area $A_{\min} = 36$, margin ratio $\alpha = 0.15$
\Ensure Cell patches $C$, bounding box positions $P$
\State Extract grayscale image $G$ from $I$
\State Apply Canny edge detection on $G$ with thresholds $(\tau_{\text{low}}, \tau_{\text{high}})$
\State Perform morphological closing using elliptical kernel of size $k$
\State Extract external contours from processed image
\State Initialize $C \leftarrow \emptyset$, $P \leftarrow \emptyset$
\For{each contour $c$}
    \State Compute contour area $A$
    \If{$A < A_{\min}$}
        \State \textbf{continue}
    \EndIf
    \State Compute bounding box $(x,y,w,h)$
    \State Expand bounding box by $\alpha$ in both dimensions
    \State Extract image patch from $G$
    \State Resize patch to $32 \times 32$
    \State Append patch to $C$ and bounding box to $P$
\EndFor
\State\Return $C$, $P$
\end{algorithmic}
\end{algorithm}

\subsection{Cellpose-SAM\texorpdfstring{~\cite{Cellpose-SAM:25}}{}}

\label{cellpose}
We evaluate Cellpose-SAM~\cite{Cellpose-SAM:25} in a zero-shot setting, i.e., without fine-tuning or retraining on our dataset, using the \texttt{CellposeModel} class with GPU acceleration and the default pretrained model, \texttt{cpsam\_v2}. Cell diameter was estimated automatically (\texttt{diameter=None}), and default flow-error (\texttt{0.4}) and cell-probability (\texttt{0.0}) thresholds were used with automatic channel detection and Cellpose's default percentile-based normalization. Predicted masks were relabeled via connected-component analysis, and axis-aligned bounding boxes were derived from each region's extent using scikit-image's \texttt{regionprops} function.

\subsection{Trained Model}
\label{Trained Model}

We train three object detectors on the same data, following the pipeline illustrated in Figure~\ref{fig:training_cell_detection}. Each brightfield image ($2080\times2784$) is divided into 100 overlapping patches of size $260\times348$ (25\% overlap), yielding 4500 patches in total, all of which are used for training. Only the first image channel is used as input. Input patches are normalized using the dataset mean ($0.810$) and standard deviation ($0.071$). Bounding boxes provided by the PLX system serve as reference annotations, and all models are trained for single-class detection (cell vs.\ background).

\subsubsection{Faster R-CNN\texorpdfstring{~\cite{FasterRCNN:15}}{}}
\begin{figure}[t]
    \centering
    \includegraphics[width=\linewidth]{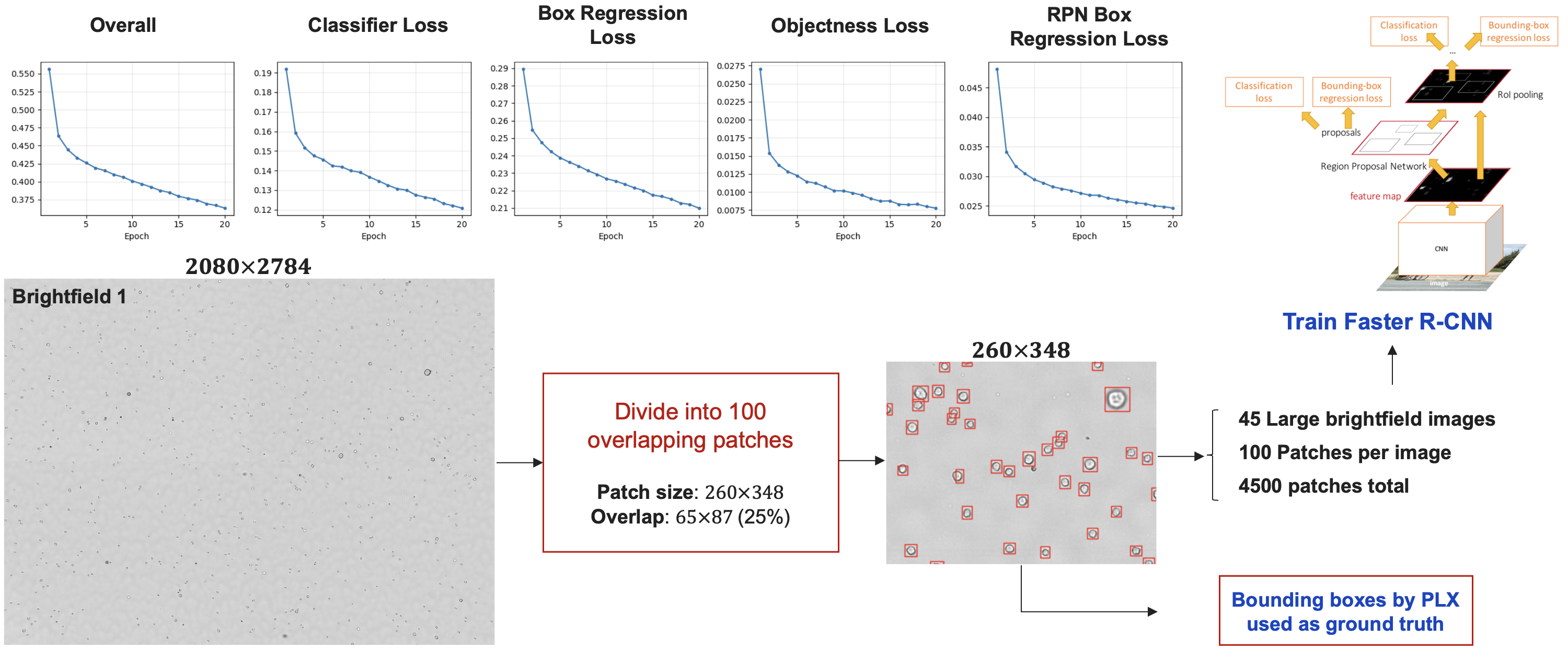}
    \caption{Training pipeline for the cell detection models. Each brightfield image is divided into 100 overlapping patches, yielding 4500 patches in total, all of which are used for training. Three detectors (Faster R-CNN, YOLO26, and RF-DETR) are trained using PLX bounding boxes as reference annotations. Training loss curves for each loss component of Faster R-CNN are shown over 20 epochs.}
    \label{fig:training_cell_detection}
\end{figure}

We fine-tune a Faster R-CNN detector~\cite{FasterRCNN:15} with a ResNet-50-FPN backbone~\cite{ResNet:15} pretrained on COCO~\cite{coco:15}, which contains approximately 41.1 million trainable parameters. The final classification head is replaced to output two classes (cell and background). The total training loss is the sum of four components: a classification loss and a bounding-box regression loss from the RoI head, and an objectness loss and a bounding-box regression loss from the Region Proposal Network (RPN). The model is optimized using SGD with a constant learning rate of $0.001$, momentum of $0.9$, weight decay of $5\times10^{-4}$, and a batch size of 4. Training is conducted for 20 epochs, during which all loss components decrease steadily, as shown in Figure~\ref{fig:training_cell_detection}.

\subsubsection{YOLO26\texorpdfstring{~\cite{yolo26:26}}{}}
We train a YOLO26-S detector~\cite{yolo26:26} using the Ultralytics framework, which contains approximately 9.9 million trainable parameters. Unlike Faster R-CNN, which is a two-stage detector that first generates region proposals with an RPN and then classifies and refines each proposal in an RoI head, YOLO26 is a single-stage, anchor-free detector that predicts bounding boxes and class scores directly from the feature maps in a single forward pass. The model is initialized from COCO-pretrained weights~\cite{coco:15}, with the classification head replaced to output a single class (cell). The PLX annotations are converted to the YOLO format: each bounding box is clipped to the patch boundaries, boxes with zero width or height are discarded, and the remaining boxes are encoded as normalized center coordinates, width, and height. Input patches are resized to $1088\times1088$. The total training loss is the sum of three components: an IoU-based bounding-box loss, a classification loss, and an L1 bounding-box regression loss. The model is optimized using AdamW~\cite{AdamW:19} with an initial learning rate of $0.002$, momentum of $0.9$, and weight decay of $5\times10^{-4}$, with a 3-epoch warm-up followed by a linear decay of the learning rate to 1\% of its initial value. Training is conducted for 20 epochs with a batch size of 16.

\subsubsection{RF-DETR\texorpdfstring{~\cite{RF-DETR:26}}{}}
We train an RF-DETR Large detector~\cite{RF-DETR:26}, a real-time transformer-based detector built on a DINOv2-based vision transformer backbone~\cite{DINOv2:24}, which contains approximately 33.6 million trainable parameters. The model is initialized from COCO-pretrained RF-DETR weights~\cite{coco:15}, with the detection head re-initialized to output a single class (cell), and is fine-tuned using the same converted annotations as YOLO26. Input patches are resized to $704\times704$. The training loss combines a classification term with L1 and generalized IoU (GIoU) bounding-box regression terms. The model is trained for 20 epochs with a batch size of 4, a learning rate of $10^{-4}$, and a random seed of 42. An exponential moving average (EMA) of the model weights is maintained during training, and the EMA weights from the final epoch are used for evaluation.

\subsection{Additional Results}
\label{Additional Results}
\subsubsection{Qualitative Comparison}
\begin{figure}[t]
    \centering
    \includegraphics[width=\linewidth]{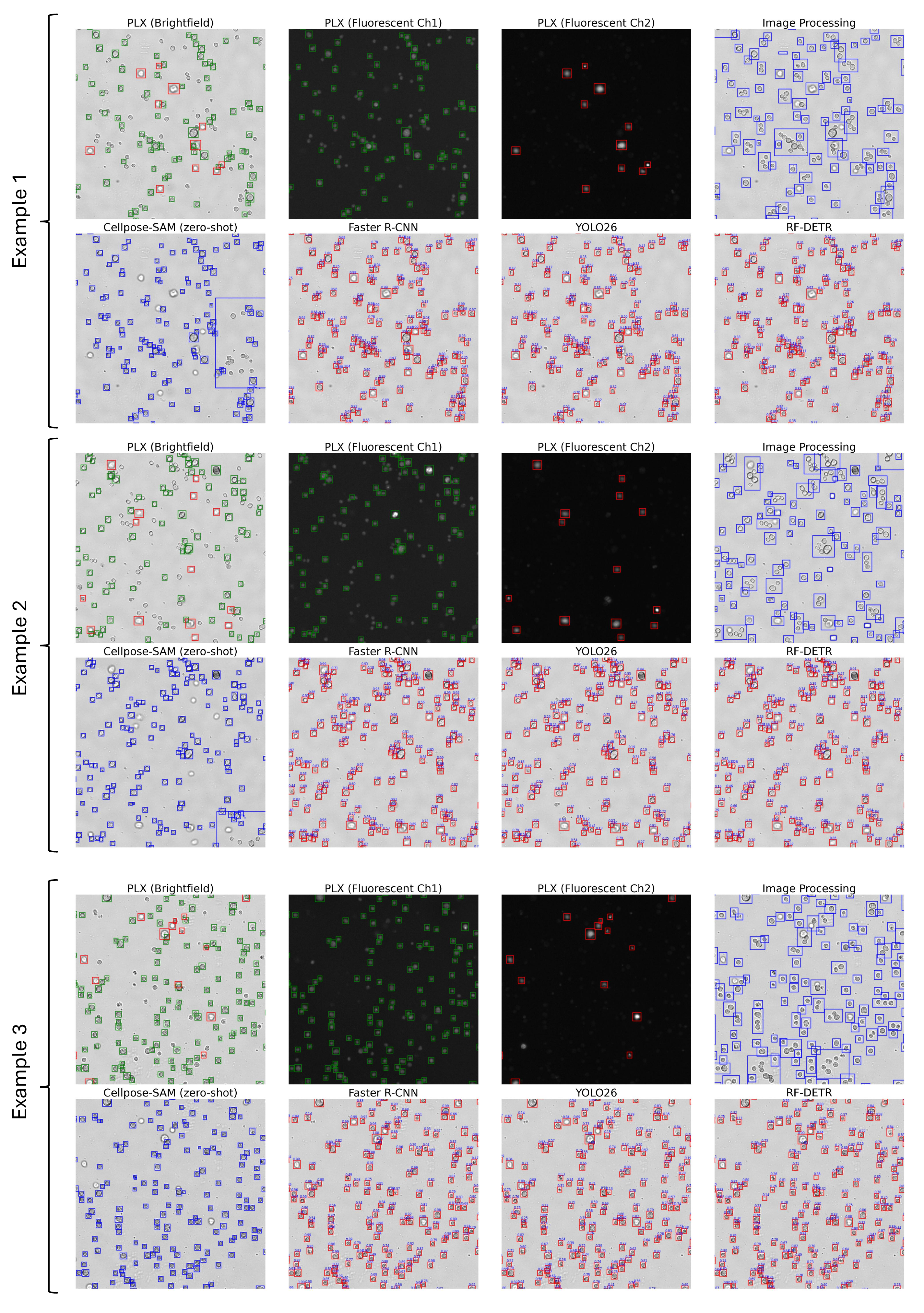}
    \caption{Qualitative comparison of cell detection methods on three \textbf{stained} brightfield images. For each example, the top row shows the PLX detections used as reference annotations, overlaid on the brightfield image (live cells in green, dead cells in red) and on the two fluorescent channels, Ch1 (live cells, green boxes) and Ch2 (dead cells, red boxes), followed by detections from the image-processing baseline (blue boxes). The bottom row shows detections from zero-shot Cellpose-SAM (blue boxes) and from the three trained models, Faster R-CNN, YOLO26, and RF-DETR (red boxes). The number above each box of the trained models is the predicted probability that the proposal is a cell.}
    \label{fig:segmentation_comparison_stained}
\end{figure}

\begin{figure}[t]
    \centering
    \includegraphics[width=\linewidth]{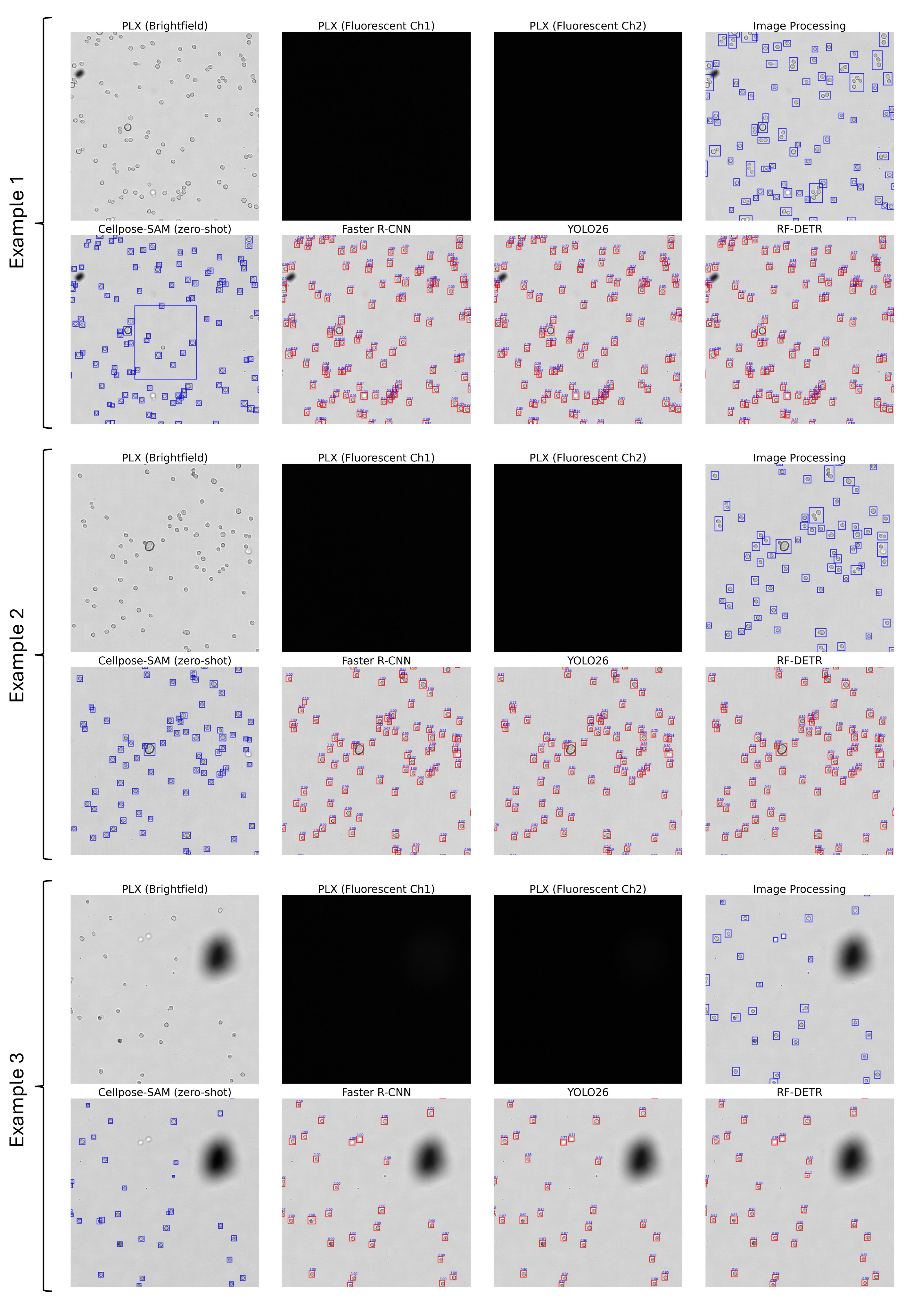}
    \caption{Qualitative comparison of cell detection methods on three \textbf{unstained} brightfield images. For each example, the top row shows the brightfield image and the two fluorescent channels, Ch1 and Ch2, which contain no signal because the samples are unstained; as a result, no PLX reference annotations are available. The top row also shows detections from the image-processing baseline (blue boxes). The bottom row shows detections from zero-shot Cellpose-SAM (blue boxes) and from the three trained models, Faster R-CNN, YOLO26, and RF-DETR (red boxes). The number above each box of the trained models is the predicted probability that the proposal is a cell.}
\label{fig:segmentation_comparison_unstained}
\end{figure}

Figures~\ref{fig:segmentation_comparison_stained} and~\ref{fig:segmentation_comparison_unstained} show additional qualitative comparisons for stained and unstained conditions, respectively. In the stained examples (Figure~\ref{fig:segmentation_comparison_stained}), the PLX system misses a notable number of cells that are nonetheless visible in the fluorescent channels, confirming these omissions stem from detection limitations rather than biological absence. The image-processing baseline performs reasonably on well-separated cells but fails to resolve clustered or overlapping cells, producing large merged bounding boxes. Zero-shot Cellpose-SAM~\cite{Cellpose-SAM:25} frequently struggles with dead cells (white appearance), generating either oversized merged regions or spurious small detections. Our trained models consistently detect the majority of cells across both isolated and clustered configurations, with confidence scores enabling further filtering of low-quality proposals. Similar patterns observed in the stained samples are also evident in the three unstained examples shown in Figure~\ref{fig:segmentation_comparison_unstained}. Although PLX detections are unavailable for unstained images due to the absence of fluorescent signal, the image-processing baseline and Cellpose-SAM~\cite{Cellpose-SAM:25} exhibit comparable failure modes, while our model maintains robust performance. In addition, detailed quantitative comparisons of cell detection methods, including cell counts, are provided in Tables~\ref{tab:plx_training_comparison} and~\ref{tab:plx_test_comparison}.

\begin{table*}[t]
\centering
\scriptsize
\setlength{\tabcolsep}{1pt}
\caption{Cell counts and viability (\%) on the 45 stained \textbf{training} images. Live/dead labels come from fluorescence for PLX and from the CNN viability classifier for the trained detectors. Image processing and Cellpose-SAM report total counts only.}
\label{tab:plx_training_comparison}
\begin{tabular}{@{}l rrr c c rrr rrr rrr@{}}
\toprule
& \multicolumn{3}{c}{\textbf{PLX}}
& \textbf{Image Proc.} & \textbf{Cellpose-SAM}
& \multicolumn{3}{c}{\textbf{Faster R-CNN}}
& \multicolumn{3}{c}{\textbf{YOLO26}}
& \multicolumn{3}{c}{\textbf{RF-DETR}} \\
\cmidrule(lr){2-4} \cmidrule(lr){5-5} \cmidrule(lr){6-6}
\cmidrule(lr){7-9} \cmidrule(lr){10-12} \cmidrule(l){13-15}
\textbf{Experiment}
  & Live & Dead & Via.\,(\%)
  & Count & Count
  & Live & Dead & Via.\,(\%)
  & Live & Dead & Via.\,(\%)
  & Live & Dead & Via.\,(\%) \\
\midrule
NIST d0--D2 & 352 & 18 & 95.1 & 367 & 422 & 341 & 25 & 93.2 & 336 & 26 & 92.8 & 341 & 25 & 93.2 \\
\addlinespace[2pt]
NIST d1--E2 & 564 & 31 & 94.8 & 573 & 591 & 550 & 40 & 93.2 & 544 & 38 & 93.5 & 550 & 46 & 92.3 \\
NIST d1--F2 & 548 & 35 & 94.0 & 568 & 586 & 533 & 45 & 92.2 & 535 & 41 & 92.9 & 533 & 49 & 91.6 \\
NIST d1--G2 & 587 & 15 & 97.5 & 616 & 624 & 581 & 16 & 97.3 & 576 & 18 & 97.0 & 576 & 24 & 96.0 \\
NIST d1--H2 & 619 & 30 & 95.4 & 614 & 654 & 601 & 35 & 94.5 & 598 & 40 & 93.7 & 598 & 38 & 94.0 \\
\addlinespace[2pt]
NIST d2--A3 & 747 & 49 & 93.8 & 820 & 812 & 730 & 71 & 91.1 & 715 & 80 & 89.9 & 718 & 90 & 88.9 \\
NIST d2--B3 & 893 & 65 & 93.2 & 907 & 924 & 862 & 72 & 92.3 & 859 & 69 & 92.6 & 860 & 78 & 91.7 \\
NIST d2--C3 & 854 & 48 & 94.7 & 842 & 899 & 805 & 72 & 91.8 & 805 & 62 & 92.8 & 812 & 73 & 91.8 \\
NIST d2--D3 & 758 & 49 & 93.9 & 764 & 816 & 743 & 56 & 93.0 & 737 & 55 & 93.1 & 741 & 64 & 92.0 \\
\addlinespace[2pt]
NIST d3--E3 & 1,568 & 92 & 94.5 & 1,533 & 1,660 & 1,532 & 115 & 93.0 & 1,529 & 102 & 93.7 & 1,535 & 119 & 92.8 \\
NIST d3--F3 & 1,934 & 113 & 94.5 & 1,876 & 2,195 & 1,904 & 119 & 94.1 & 1,905 & 118 & 94.2 & 1,909 & 127 & 93.8 \\
NIST d3--G3 & 1,556 & 80 & 95.1 & 1,518 & 1,709 & 1,511 & 101 & 93.7 & 1,506 & 101 & 93.7 & 1,522 & 109 & 93.3 \\
NIST d3--H3 & 1,432 & 92 & 94.0 & 1,376 & 1,533 & 1,395 & 93 & 93.8 & 1,386 & 85 & 94.2 & 1,399 & 96 & 93.6 \\
\addlinespace[2pt]
NIST d4--A1 & 2,354 & 135 & 94.6 & 2,205 & 2,589 & 2,303 & 155 & 93.7 & 2,308 & 144 & 94.1 & 2,320 & 167 & 93.3 \\
NIST d4--B1 & 2,589 & 136 & 95.0 & 2,432 & 2,718 & 2,525 & 167 & 93.8 & 2,519 & 155 & 94.2 & 2,533 & 192 & 93.0 \\
NIST d4--C1 & 2,431 & 155 & 94.0 & 2,221 & 2,579 & 2,378 & 146 & 94.2 & 2,372 & 139 & 94.5 & 2,385 & 167 & 93.5 \\
NIST d4--D1 & 2,088 & 125 & 94.4 & 1,972 & 2,360 & 2,056 & 137 & 93.8 & 2,058 & 124 & 94.3 & 2,065 & 148 & 93.3 \\
\addlinespace[2pt]
NIST d5--E1 & 3,477 & 216 & 94.2 & 2,980 & 3,777 & 3,423 & 227 & 93.8 & 3,416 & 202 & 94.4 & 3,440 & 269 & 92.7 \\
NIST d5--F1 & 4,868 & 246 & 95.2 & 3,659 & 5,504 & 4,767 & 277 & 94.5 & 4,788 & 277 & 94.5 & 4,828 & 324 & 93.7 \\
NIST d5--G1 & 3,789 & 191 & 95.2 & 3,199 & 4,409 & 3,721 & 183 & 95.3 & 3,720 & 167 & 95.7 & 3,744 & 205 & 94.8 \\
NIST d5--H1 & 3,617 & 193 & 94.9 & 3,090 & 4,368 & 3,555 & 176 & 95.3 & 3,553 & 172 & 95.4 & 3,581 & 190 & 95.0 \\
\addlinespace[2pt]
NIST d6--A2 & 2,804 & 140 & 95.2 & 2,391 & 3,117 & 2,839 & 180 & 94.0 & 2,832 & 161 & 94.6 & 2,858 & 211 & 93.1 \\
NIST d6--B2 & 3,373 & 210 & 94.1 & 2,800 & 3,884 & 3,384 & 235 & 93.5 & 3,376 & 225 & 93.8 & 3,411 & 274 & 92.6 \\
NIST d6--C2 & 2,649 & 141 & 94.9 & 2,375 & 2,871 & 2,621 & 177 & 93.7 & 2,620 & 169 & 93.9 & 2,635 & 185 & 93.4 \\
NIST d6--D2 & 2,530 & 127 & 95.2 & 2,279 & 2,708 & 2,468 & 147 & 94.4 & 2,463 & 135 & 94.8 & 2,484 & 158 & 94.0 \\
\addlinespace[2pt]
NIST d7--AFBa & 1,648 & 227 & 87.9 & 2,499 & 3,608 & 3,277 & 333 & 90.8 & 3,225 & 304 & 91.4 & 3,286 & 353 & 90.3 \\
NIST d7--AFBb & 1,980 & 286 & 87.4 & 2,564 & 3,774 & 3,411 & 356 & 90.5 & 3,394 & 328 & 91.2 & 3,427 & 366 & 90.4 \\
NIST d7--FUSa & 2,610 & 216 & 92.4 & 2,375 & 3,856 & 3,287 & 258 & 92.7 & 3,280 & 240 & 93.2 & 3,316 & 264 & 92.6 \\
NIST d7--FUSb & 2,417 & 236 & 91.1 & 2,719 & 3,489 & 3,162 & 262 & 92.3 & 3,159 & 264 & 92.3 & 3,177 & 291 & 91.6 \\
\addlinespace[2pt]
NIST d8--AFBa & 3,183 & 271 & 92.2 & 3,105 & 3,807 & 3,373 & 421 & 88.9 & 3,368 & 373 & 90.0 & 3,403 & 494 & 87.3 \\
NIST d8--AFBb & 3,138 & 337 & 90.3 & 3,273 & 4,016 & 3,494 & 492 & 87.7 & 3,506 & 435 & 89.0 & 3,525 & 596 & 85.5 \\
NIST d8--FUSa & 3,467 & 289 & 92.3 & 3,378 & 4,134 & 3,804 & 369 & 91.2 & 3,797 & 346 & 91.6 & 3,825 & 412 & 90.3 \\
NIST d8--FUSb & 3,285 & 270 & 92.4 & 3,145 & 3,844 & 3,472 & 328 & 91.4 & 3,456 & 317 & 91.6 & 3,484 & 378 & 90.2 \\
\addlinespace[2pt]
NIST d9--AFBa & 1,859 & 301 & 86.1 & 2,631 & 3,945 & 3,073 & 757 & 80.2 & 2,964 & 580 & 83.6 & 3,072 & 763 & 80.1 \\
NIST d9--AFBb & 2,731 & 476 & 85.2 & 3,463 & 4,561 & 3,810 & 757 & 83.4 & 3,825 & 666 & 85.2 & 3,863 & 900 & 81.1 \\
NIST d9--FUSa & 2,919 & 301 & 90.7 & 3,012 & 4,152 & 3,581 & 493 & 87.9 & 3,589 & 424 & 89.4 & 3,601 & 529 & 87.2 \\
NIST d9--FUSb & 2,992 & 297 & 91.0 & 3,507 & 4,372 & 3,835 & 466 & 89.2 & 3,853 & 432 & 89.9 & 3,865 & 519 & 88.2 \\
\addlinespace[2pt]
NIST d10--AFBa & 2,459 & 392 & 86.3 & 3,136 & 4,186 & 3,401 & 734 & 82.2 & 3,398 & 631 & 84.3 & 3,437 & 847 & 80.2 \\
NIST d10--AFBb & 2,508 & 778 & 76.3 & 3,540 & 4,656 & 3,692 & 971 & 79.2 & 3,668 & 831 & 81.5 & 3,722 & 1,450 & 72.0 \\
NIST d10--FUSa & 3,356 & 423 & 88.8 & 3,414 & 4,601 & 3,849 & 744 & 83.8 & 3,858 & 648 & 85.6 & 3,910 & 799 & 83.0 \\
NIST d10--FUSb & 3,221 & 382 & 89.4 & 3,322 & 4,444 & 3,837 & 570 & 87.1 & 3,843 & 494 & 88.6 & 3,861 & 658 & 85.4 \\
\addlinespace[2pt]
NIST d13--AFBa & 2,431 & 721 & 77.1 & 3,129 & 4,035 & 3,193 & 818 & 79.6 & 3,183 & 681 & 82.4 & 3,194 & 1,314 & 70.9 \\
NIST d13--AFBb & 1,919 & 1,194 & 61.6 & 2,849 & 4,203 & 2,860 & 1,267 & 69.3 & 2,857 & 1,167 & 71.0 & 2,845 & 1,559 & 64.6 \\
NIST d13--FUSa & 1,766 & 906 & 66.1 & 2,963 & 4,294 & 3,093 & 1,151 & 72.9 & 3,105 & 1,043 & 74.9 & 3,100 & 1,451 & 68.1 \\
NIST d13--FUSb & 2,495 & 547 & 82.0 & 3,189 & 4,048 & 3,012 & 852 & 78.0 & 3,012 & 716 & 80.8 & 3,047 & 1,418 & 68.2 \\
\midrule
\textbf{Total} & 101,365 & 11,582 & $-$ & 105,190 & 136,334 & 116,644 & 15,466 & $-$ & 116,396 & 13,825 & $-$ & 117,338 & 18,789 & $-$ \\
\bottomrule
\end{tabular}
\end{table*}

\begin{table*}[t]
\centering
\scriptsize
\setlength{\tabcolsep}{1pt}
\caption{Cell counts and viability (\%) on the stained and unstained \textbf{test} images. Live/dead labels come from fluorescence for PLX and from the CNN viability classifier for the trained detectors. Image processing and Cellpose-SAM report total counts only. PLX annotations are unavailable for unstained images (--).}
\label{tab:plx_test_comparison}
\begin{tabular}{@{}l rrr c c rrr rrr rrr@{}}
\toprule
& \multicolumn{3}{c}{\textbf{PLX}}
& \textbf{Image Proc.} & \textbf{Cellpose-SAM}
& \multicolumn{3}{c}{\textbf{Faster R-CNN}}
& \multicolumn{3}{c}{\textbf{YOLO26}}
& \multicolumn{3}{c}{\textbf{RF-DETR}} \\
\cmidrule(lr){2-4} \cmidrule(lr){5-5} \cmidrule(lr){6-6}
\cmidrule(lr){7-9} \cmidrule(lr){10-12} \cmidrule(l){13-15}
\textbf{Experiment}
  & Live & Dead & Via.\,(\%)
  & Count & Count
  & Live & Dead & Via.\,(\%)
  & Live & Dead & Via.\,(\%)
  & Live & Dead & Via.\,(\%) \\
\midrule
70\%-Stained-C1 & 372 & 146 & 71.8 & 508 & 624 & 425 & 159 & 72.8 & 411 & 187 & 68.7 & 415 & 164 & 71.7 \\
70\%-Stained-C2 & 360 & 133 & 73.0 & 500 & 523 & 403 & 149 & 73.0 & 388 & 168 & 69.8 & 390 & 152 & 72.0 \\
70\%-Stained-C3 & 352 & 163 & 68.3 & 505 & 599 & 376 & 182 & 67.4 & 382 & 176 & 68.5 & 375 & 169 & 68.9 \\
\addlinespace[2pt]
80\%-Stained-A1 & 308 & 72 & 81.1 & 410 & 411 & 321 & 100 & 76.2 & 299 & 118 & 71.7 & 316 & 108 & 74.5 \\
80\%-Stained-A2 & 297 & 76 & 79.6 & 383 & 335 & 325 & 79 & 80.4 & 321 & 87 & 78.7 & 321 & 85 & 79.1 \\
80\%-Stained-A3 & 322 & 84 & 79.3 & 381 & 380 & 318 & 98 & 76.4 & 310 & 112 & 73.5 & 309 & 107 & 74.3 \\
\addlinespace[2pt]
87\%-Stained-F1 & 336 & 39 & 89.6 & 373 & 362 & 329 & 54 & 85.9 & 323 & 54 & 85.7 & 318 & 61 & 83.9 \\
87\%-Stained-F2 & 306 & 55 & 84.8 & 350 & 365 & 299 & 68 & 81.5 & 298 & 68 & 81.4 & 298 & 71 & 80.8 \\
87\%-Stained-F3 & 334 & 55 & 85.9 & 375 & 400 & 329 & 63 & 83.9 & 325 & 58 & 84.9 & 322 & 60 & 84.3 \\
\addlinespace[2pt]
95\%-Stained-A1 & 310 & 14 & 95.7 & 312 & 292 & 291 & 25 & 92.1 & 293 & 23 & 92.7 & 291 & 24 & 92.4 \\
95\%-Stained-A2 & 272 & 11 & 96.1 & 283 & 272 & 266 & 17 & 94.0 & 264 & 13 & 95.3 & 262 & 15 & 94.6 \\
95\%-Stained-A3 & 319 & 22 & 93.5 & 314 & 325 & 305 & 25 & 92.4 & 300 & 24 & 92.6 & 298 & 22 & 93.1 \\
\addlinespace[2pt]
97\%-Stained-A1 & 1,239 & 36 & 97.2 & 1,148 & 1,283 & 1,192 & 50 & 96.0 & 1,192 & 47 & 96.2 & 1,199 & 50 & 96.0 \\
97\%-Stained-A2 & 908 & 23 & 97.5 & 856 & 993 & 879 & 43 & 95.3 & 817 & 109 & 88.2 & 854 & 71 & 92.3 \\
97\%-Stained-A3 & 1,047 & 45 & 95.9 & 944 & 1,140 & 1,015 & 71 & 93.5 & 1,021 & 59 & 94.5 & 1,023 & 68 & 93.8 \\
\midrule
\textbf{Total (stained)} & 7,082 & 974 & $-$ & 7,642 & 8,304 & 7,073 & 1,183 & $-$ & 6,944 & 1,303 & $-$ & 6,991 & 1,227 & $-$ \\
\midrule
70\%-Unstained-R1 & -- & -- & -- & 865 & 1,083 & 757 & 222 & 77.3 & 716 & 268 & 72.8 & 737 & 243 & 75.2 \\
70\%-Unstained-R2 & -- & -- & -- & 832 & 1,061 & 782 & 186 & 80.8 & 791 & 208 & 79.2 & 770 & 211 & 78.5 \\
70\%-Unstained-R3 & -- & -- & -- & 847 & 1,062 & 781 & 197 & 79.9 & 781 & 203 & 79.4 & 764 & 205 & 78.8 \\
\addlinespace[2pt]
80\%-Unstained-R1 & -- & -- & -- & 794 & 905 & 684 & 168 & 80.3 & 694 & 157 & 81.6 & 688 & 157 & 81.4 \\
80\%-Unstained-R2 & -- & -- & -- & 705 & 814 & 621 & 130 & 82.7 & 624 & 135 & 82.2 & 616 & 140 & 81.5 \\
80\%-Unstained-R3 & -- & -- & -- & 781 & 870 & 660 & 164 & 80.1 & 659 & 176 & 78.9 & 660 & 169 & 79.6 \\
\addlinespace[2pt]
87\%-Unstained-R1 & -- & -- & -- & 775 & 887 & 734 & 92 & 88.9 & 739 & 94 & 88.7 & 737 & 100 & 88.1 \\
87\%-Unstained-R2 & -- & -- & -- & 662 & 736 & 623 & 88 & 87.6 & 634 & 80 & 88.8 & 624 & 86 & 87.9 \\
87\%-Unstained-R3 & -- & -- & -- & 712 & 807 & 664 & 96 & 87.4 & 670 & 94 & 87.7 & 675 & 90 & 88.2 \\
\addlinespace[2pt]
95\%-Unstained-R1 & -- & -- & -- & 607 & 694 & 632 & 31 & 95.3 & 635 & 23 & 96.5 & 623 & 34 & 94.8 \\
95\%-Unstained-R2 & -- & -- & -- & 608 & 682 & 625 & 42 & 93.7 & 633 & 32 & 95.2 & 624 & 36 & 94.5 \\
95\%-Unstained-R3 & -- & -- & -- & 645 & 743 & 677 & 37 & 94.8 & 680 & 33 & 95.4 & 675 & 39 & 94.5 \\
\addlinespace[2pt]
97\%-Unstained-R1 & -- & -- & -- & 1,911 & 2,504 & 2,343 & 38 & 98.4 & 2,353 & 35 & 98.5 & 2,363 & 52 & 97.8 \\
97\%-Unstained-R2 & -- & -- & -- & 1,813 & 2,484 & 2,273 & 56 & 97.6 & 2,261 & 55 & 97.6 & 2,299 & 63 & 97.3 \\
97\%-Unstained-R3 & -- & -- & -- & 1,748 & 2,335 & 2,140 & 54 & 97.5 & 2,153 & 40 & 98.2 & 2,161 & 52 & 97.7 \\
\midrule
\textbf{Total (unstained)} & -- & -- & -- & 14,305 & 17,667 & 14,996 & 1,601 & $-$ & 15,023 & 1,633 & $-$ & 15,016 & 1,677 & $-$ \\
\bottomrule
\end{tabular}
\end{table*}

\subsubsection{Robustness to Different Focal Planes}
\begin{figure}[t]
    \centering
    \includegraphics[width=\linewidth]{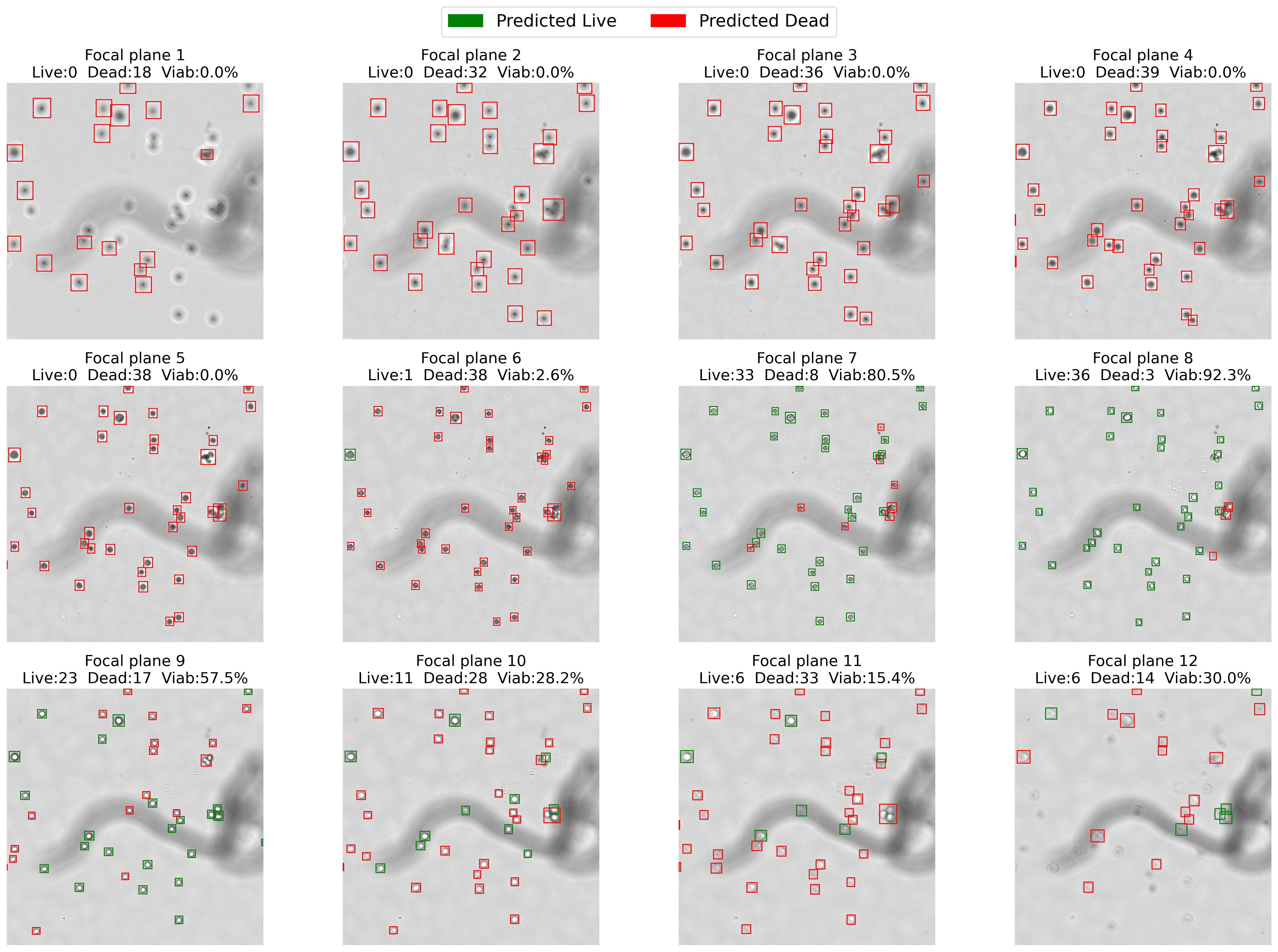}
       \caption{Performance of our pipeline on cells at twelve different focal planes. Cells are also moving, so frames are not perfectly registrable. A video of the focal stack is available in the \href{https://amirrezavazifeh.github.io/ViabiLens-Project-Page/}{project page}.}   
    \label{fig:focal_stack_supp}
\end{figure}

Figure~\ref{fig:focal_stack_supp} illustrates the performance of our pipeline across a range of focal planes. Cell appearance varies substantially with focal depth, which affects both detection and viability classification. Our cell detector remains robust across the focal planes 3 to 11, successfully localizing the majority of cells despite changes in appearance. However, viability predictions vary considerably when cells are out of focus (focal planes other than 8), as these fall outside the training distribution of the viability classifier. This suggests that while our detection model generalizes reasonably well across focal planes, the viability classifier performs most reliably near the focal plane used during training. 

\subsubsection{Quantitative Results}
Tables~\ref{tab:plx_training_comparison} and~\ref{tab:plx_test_comparison} report the detailed results corresponding to the figures in the main paper, including the total cell count and overall viability prediction for each method. Table~\ref{tab:plx_training_comparison} presents the results of applying our pipeline to the training images, which contain only stained samples. Table~\ref{tab:plx_test_comparison} presents the results on the test set, which includes both stained and unstained samples, and corresponds to Figs.~\ref{fig:viability_result_stained} and~\ref{fig:viability_result_unstained}.

\section{Evaluation Details}
\label{Evaluation Details}
We define below the metrics used to evaluate cell detection, live/dead classification, and end-to-end viability estimation.

\subsection{Cell Detection}
\subsubsection{Precision, Recall, and F1 Score}
A predicted bounding box is a true positive (TP) if its Intersection over Union (IoU) with a ground-truth box exceeds a threshold, and a false positive (FP) otherwise. Ground-truth boxes with no matching prediction are false negatives (FN). IoU is defined as
\begin{equation}
\text{IoU} = \frac{\text{Area}(B_{\text{pred}} \cap B_{\text{gt}})}{\text{Area}(B_{\text{pred}} \cup B_{\text{gt}})}.
\end{equation}
Precision is the fraction of predictions that match a ground-truth cell, recall is the fraction of ground-truth cells that are detected, and the F1 score is their harmonic mean:
\begin{equation}
\text{Precision} = \frac{TP}{TP + FP}, \qquad
\text{Recall} = \frac{TP}{TP + FN}, \qquad
F_1 = \frac{2 \cdot \text{Precision} \cdot \text{Recall}}{\text{Precision} + \text{Recall}}.
\label{eq:prf}
\end{equation}

\subsubsection{Mean Average Precision at IoU$=$0.5}
Average precision (AP) is the area under the precision-recall curve obtained by ranking predictions by confidence score:
\begin{equation}
\text{AP} = \sum_{k} (r_k - r_{k-1})\, p_k,
\label{eq:ap}
\end{equation}
where $p_k$ and $r_k$ are the precision and recall at the $k$-th confidence threshold. We use an IoU threshold of 0.5, compute AP for each test image, and report the mean across images as mAP@0.5. Because cells are the only object class, the mean is taken over images rather than classes. AP is reported only for trained models, since the image-processing baseline and Cellpose-SAM (zero-shot) do not produce confidence scores.

\subsection{Cell Classification}
For live/dead classification, dead cells are the positive class and live cells the negative class. TP, FP, and FN denote dead cells correctly classified as dead, live cells misclassified as dead, and dead cells misclassified as live, respectively; TN denotes live cells correctly classified as live. Precision, recall, and F1 follow Eq.~\ref{eq:prf}, and AP follows Eq.~\ref{eq:ap}, with thresholds applied to the predicted dead-cell probability. We also report accuracy, the fraction of correctly classified cells:
\begin{equation}
\text{Accuracy} = \frac{TP + TN}{N},
\end{equation}
where $N$ is the total number of cells.

\subsection{End-to-End Viability Estimation}
We compare the viability predicted by our pipeline, $V_i^{\text{pred}}$, with the measured viability provided by PLX, $V_i^{\text{GT}}$, over $M = 15$ images. We report the mean absolute error (MAE) and the coefficient of determination ($R^2$), which measures how much of the variance in ground-truth viability is explained by the predictions:
\begin{equation}
\text{MAE} = \frac{1}{M}\sum_{i=1}^{M} \left| V_i^{\text{GT}} - V_i^{\text{pred}} \right|,
\end{equation}
\begin{equation}
R^2 = 1 - \frac{\sum_{i=1}^{M} \left(V_i^{\text{GT}} - V_i^{\text{pred}}\right)^2}{\sum_{i=1}^{M} \left(V_i^{\text{GT}} - \bar{V}^{\text{GT}}\right)^2},
\end{equation}
where $\bar{V}^{\text{GT}}$ is the mean ground-truth viability.

\section{Zeiss Microscope}
\label{ZEISS Microscope}

To confirm the generalizability of our results, we replaced the PLX image cytometer with a high-resolution Zeiss microscope. The processing details and corresponding results for this dataset are provided below.

\subsection{Sample Preparation}
NISTCHO, a clonal CHO-K1 cell line producing cNISTmAb (NIST, SKU 8675), was thawed and cultured in 30~mL of EX-CELL\textregistered{} CD CHO Fusion medium (MilliporeSigma, cat.~14365C) in a 125~mL shake flask at 37~$^\circ$C, 135~rpm (25~mm throw), 80\% relative humidity, and 5\% CO$_2$. Cells were passaged every 3--4 days by dilution into fresh medium and seeded at 0.3--0.4~$\times 10^{6}$ cells/mL. To obtain samples spanning the range of viabilities typically encountered in bioprocessing, two fed-batch cultures, designated V1 and V2, were inoculated from the same seed train at 0.5~$\times10^{6}$~cells/mL in EX-CELL\textsuperscript{\textregistered} Advanced CHO Fed-batch Medium (MilliporeSigma, cat.~14365C) in two separate vessels of an Ambr\textsuperscript{\textregistered}~250 bioreactor system (Sartorius), each with an initial working volume of 205~mL, and run in parallel under identical process conditions. pH was set to 7.0 with a deadband of $\pm$0.1 and controlled by a PID loop via CO$_2$ sparging and addition of 1~M sodium carbonate (MilliporeSigma, cat.~S7795). Bolus additions of Gibco FoamAway antifoaming agent (Thermo Fisher Scientific, cat.~A1036902) were made as needed to control foaming. Impeller agitation was set to 400~rpm, dissolved oxygen to 30\%, and the air sparge rate was initially 2~mL/min and increased to 5~mL/min on day~6. Starting on day~3, EX-CELL Advanced CHO Feed~1 (MilliporeSigma, cat.~24368C) was added daily at 2.5\% (v/v), and a 400~g/L glucose solution (MilliporeSigma, cat.~G7021) was fed daily to restore the glucose concentration to 6~g/L. Both cultures were maintained for 14 days, with samples taken from each vessel at multiple time points for imaging.

\subsection{Image Acquisition}
\begin{figure}[t]
    \centering
    \includegraphics[width=\linewidth]{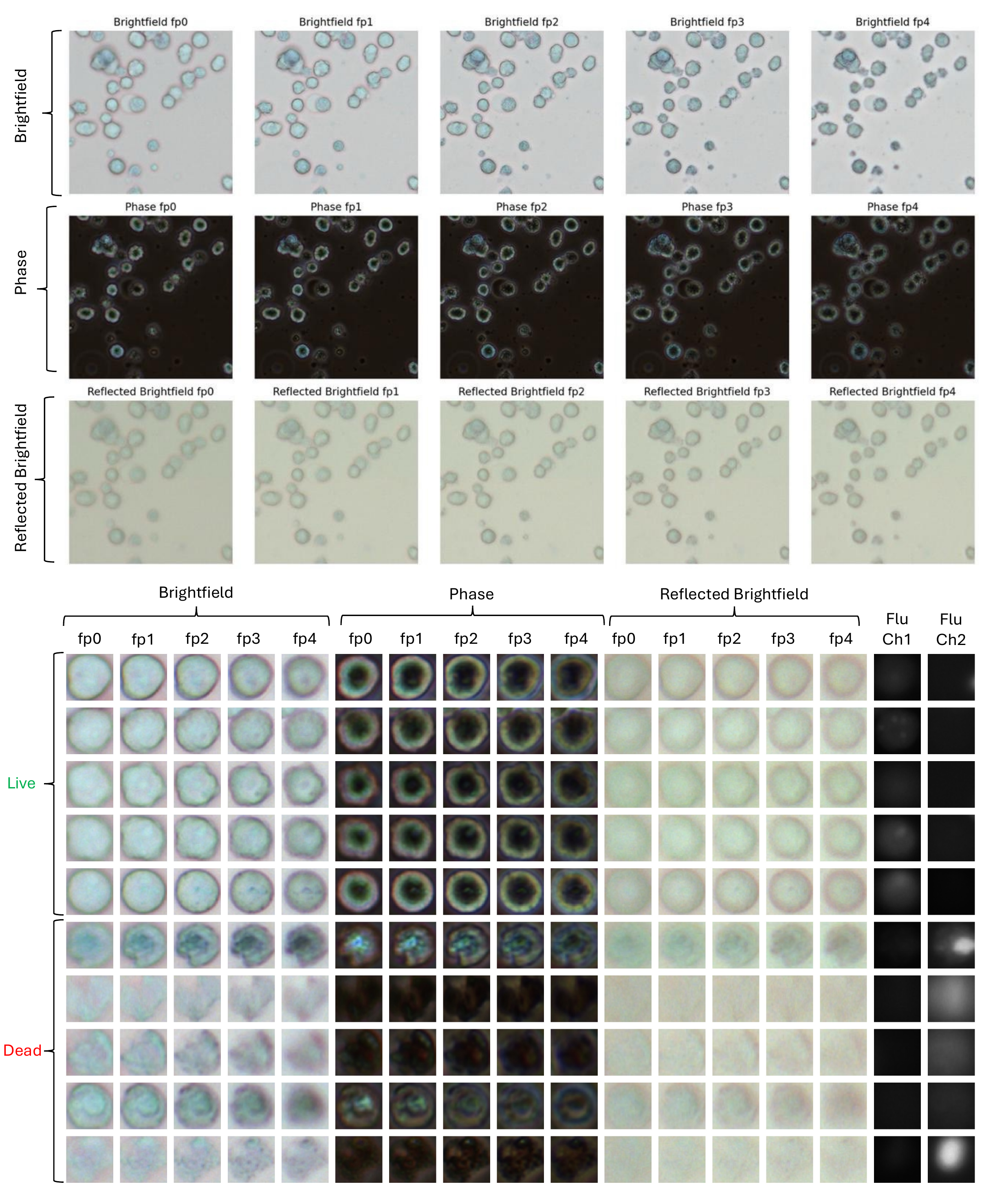}
    \caption{Representative examples from the Zeiss dataset. Top: Image crops across three modalities (brightfield, phase, and reflected brightfield) and five focal planes (fp0--fp4). Bottom: Individual cell crops across all modalities, focal planes, and fluorescent channels (Ch1, Ch2), with the first five rows showing live cells and the last five rows showing dead cells.}
    \label{fig:ziess_image_examples}
\end{figure}

Images were collected from both vessels (V1 and V2). Samples from each vessel were stained at a 1:1 volumetric ratio with ViaStain\texttrademark{} AO-PI staining solution (Revvity, part no.~CS2-0106-5ML) and incubated for 10 minutes prior to imaging. Unstained images were acquired immediately before dye addition using identical imaging settings. We acquired images at resolution 9684$\times$11604 using a Zeiss microscope (Axio Observer, EC Plan-Neofluar 10$\times$/0.30 Ph1 objective, item no.~440331-0000-000) across three modalities (brightfield, phase, and reflected brightfield). Each modality was captured at five focal planes (fp0--fp4), where the middle focal plane (fp2) is in focus. In addition, two fluorescence channels were recorded for each stained image, where live cells emit in the first channel and dead cells in the second, using Acridine Orange (AO) and Propidium Iodide (PI) dyes for viability labeling (488/509~nm excitation/emission at 258~ms exposure for AO; 548/561~nm at 1.29~s for PI). Because ground-truth viability labels for individual cells are not directly available and must instead be inferred from the fluorescence channels, we also measured the overall viability of each sample with the Cellaca\texttrademark{} PLX image cytometer, providing an independent reference for the fluorescence-derived labels. Samples were taken from both vessels at multiple time points over a two-week period, spanning the range of viabilities observed during the culture. All stained and unstained samples used in this study, along with their sampling days, are listed in Table~\ref{tab:ziess_overall_viability}. Representative examples of individual cells and image crops are shown in Figure~\ref{fig:ziess_image_examples}.

\subsection{Cell Detection}
Unlike the PLX dataset, for which we developed a dedicated AI-based cell detector, we used Cellpose-SAM \cite{Stringer:21, Cellpose-SAM:25} for the Zeiss dataset, as it provides strong detection performance. Since Cellpose-SAM may return a different number of detected cells across modalities and focal planes, we used the bounding boxes obtained by applying it to the in-focus focal plane (fp2) of brightfield images as the baseline for all other modalities and focal planes. In total, the dataset comprises 85{,}304 stained and 199{,}914 unstained individual cells, acquired across 15 configurations (3~imaging modalities $\times$ 5~focal planes). Five examples of each class (live/dead) are shown in Figure \ref{fig:ziess_image_examples}.

\subsection{Cell Labeling}
Ground-truth live/dead labels for the stained cells were derived from the two fluorescence channels. For each detected cell, we computed the average intensity, along with other intensity statistics, in the AO (Ch1) and PI (Ch2) channels within its bounding box. We then compared these values between the two channels for each cell. A cell was labeled dead if its PI signal exceeded its AO signal by more than a threshold, and live otherwise. Because there is no direct cell-level ground truth, we chose this threshold so that the overall viability of each sample, computed from the resulting labels, matched the viability measured with the PLX cytometer as closely as possible. The resulting labels were used both to train the classifier and as the reference for evaluating its predictions on stained samples.

\subsection{Cell Classification}
\begin{figure}[t]
    \centering
    \includegraphics[width=\linewidth]{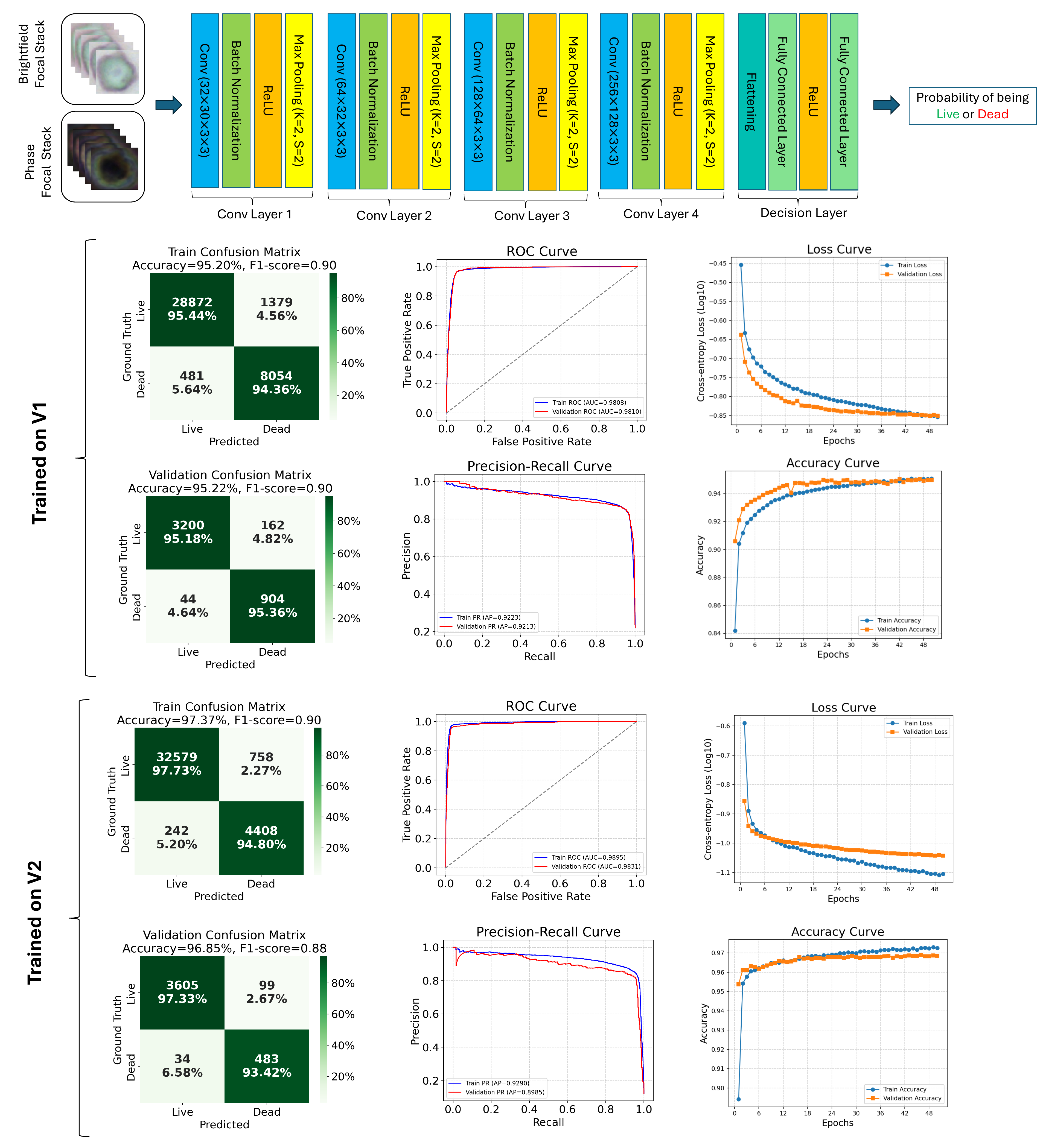}
    \caption{Live/dead classification network and its evaluation. Top: The CNN takes paired brightfield and phase images of a detected cell as input and passes them through four convolutional blocks, each consisting of convolution, batch normalization, ReLU, and max pooling, followed by a decision layer of two fully connected layers that outputs live/dead class probabilities. Bottom: training and validation confusion matrices, ROC and precision--recall curves, and accuracy and cross-entropy loss over 50 epochs, for the models trained on V1 (upper panels) and V2 (lower panels).}
    \label{fig:classification_ZIESS}
\end{figure}

For leave-one-vessel-out evaluation, we trained two classifiers with identical architecture and training settings: one on stained cells from V1 and one on stained cells from V2. Ground-truth live/dead labels for these cells were derived from the AO and PI fluorescence channels. The 43,096 V1 cells were split into 90\% training (38,786 cells) and 10\% validation (4,310 cells) sets, and the 42,208 V2 cells were split in the same way into training (37,987 cells) and validation (4,221 cells) sets. Because both subsets come from the same vessel, the validation set was used only to monitor training; model performance is evaluated on the held-out vessel (Section~\ref{sec:zeiss_eval}). The CNN model, using the architecture shown in Figure~\ref{fig:classification_ZIESS}, was trained for 50 epochs with cross-entropy loss to assign viability labels to individual cells. Unlike the PLX dataset, where only brightfield images were used as CNN input, the model for the Zeiss dataset takes brightfield and phase images from all five focal planes as input. The V1 model achieved 95.20\% training accuracy (F1-score = 0.90) and 95.22\% validation accuracy (F1-score = 0.90), and the V2 model achieved 97.37\% training accuracy (F1-score = 0.90) and 96.85\% validation accuracy (F1-score = 0.88). The confusion matrices, loss and accuracy curves, ROC curves, and precision-recall curves are shown in Figure~\ref{fig:classification_ZIESS}.

\subsection{Evaluation}
\label{sec:zeiss_eval}
We first applied our full pipeline (Cellpose-SAM for cell detection followed by CNN for viability prediction) to all 11 stained Zeiss samples, six from V1 and five from V2. For each sample, the reference viability was computed from the fluorescence-derived (AO/PI) cell labels, which we checked against independent PLX image cytometer measurements of the same samples (Table~\ref{tab:ziess_overall_viability}). We report performance on the held-out vessel only; results on the training vessel are in-sample and are listed in Table~\ref{tab:ziess_overall_viability} for completeness. The model trained on V1 achieved a mean absolute error (MAE) of $2.02\%$ on V2, and the model trained on V2 achieved an MAE of $2.82\%$ on V1.

We then applied the pipeline to the unstained images. Because cell-level ground truth is unavailable for unstained cells, we used the viability measured on the corresponding stained sample (same vessel and day) as the reference. For Day~8 V2, which has no stained Zeiss data, we used the PLX measurement of the same sample as the reference instead. This gave 12 unstained samples, six from each vessel. On the held-out vessel, the model trained on V1 achieved an MAE of $3.04\%$ on V2, and the model trained on V2 achieved an MAE of $3.84\%$ on V1.

\begin{figure}[t]
    \centering
    \includegraphics[width=\linewidth]{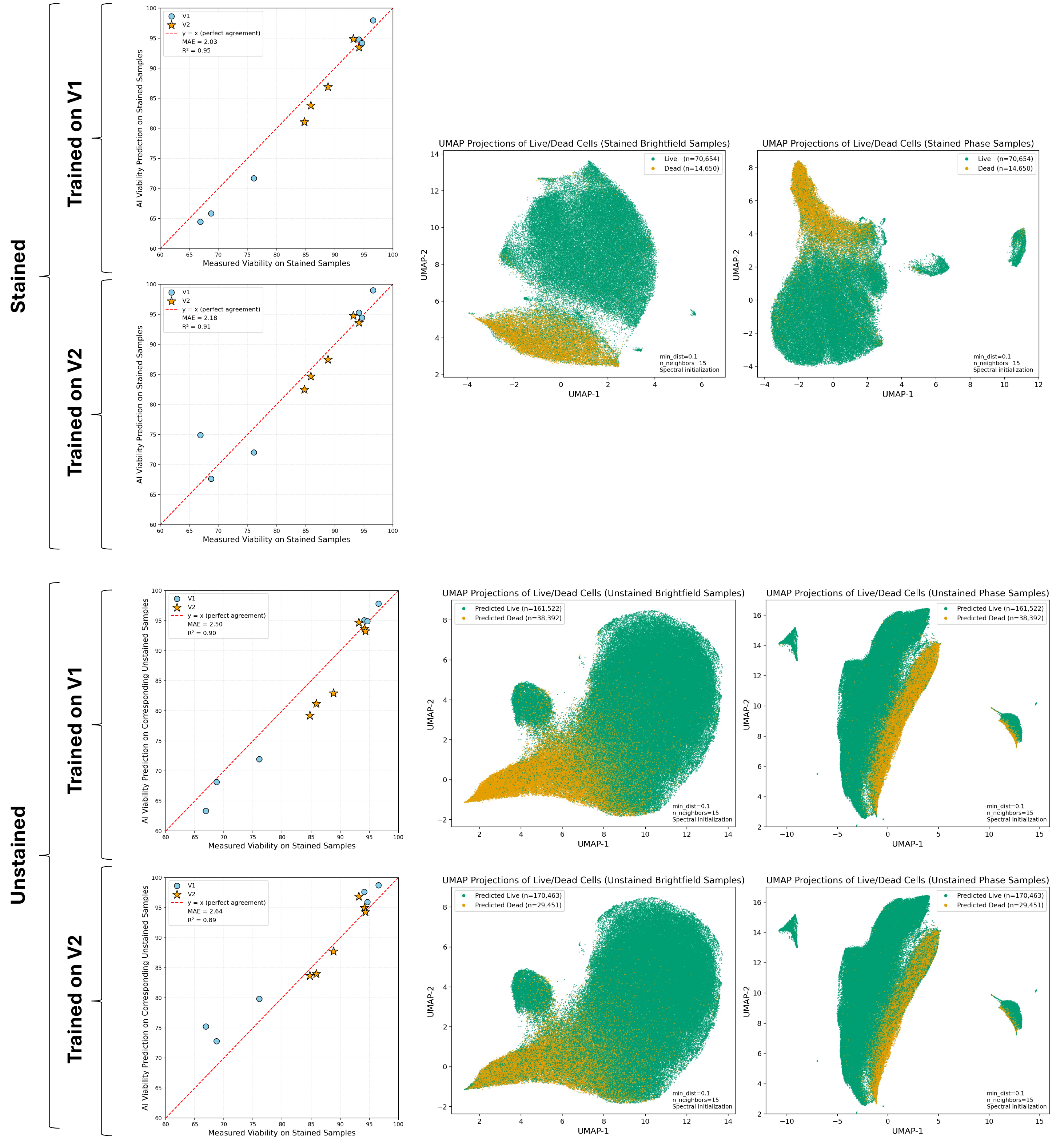}
    \caption{Evaluation results for stained (top two rows) and unstained (bottom two rows) Zeiss samples. Left: scatter plots comparing AI-predicted viability against reference measurements.  Middle and right: UMAP projections of individual cell embeddings from brightfield and phase images (middle focal plane), respectively. For stained samples, cells are colored by fluorescence-derived live/dead labels; for unstained samples, colors reflect predictions of the CNNs trained on V1 and V2.}
    \label{fig:ziess_overall_viability_result}
\end{figure}

For both stained and unstained samples, the largest deviations occurred on Days~12--14, when the viability of the two vessels diverged: V1 declined to 65--76\%, whereas V2 remained at 84--90\%. Each model was biased toward the viability distribution of its training vessel. The model trained on V1, in which about 22\% of training cells were dead, underestimated the viability of V2, whereas the model trained on V2, in which about 12\% of training cells were dead, overestimated the viability of the low-viability V1 samples (e.g., $74.9\%$ predicted vs.\ $66.9\%$ measured on Day~14). This indicates that the training data should span the range of viabilities expected at deployment. Scatter plots for the stained and unstained samples are shown in the first and second rows of Figure~\ref{fig:ziess_overall_viability_result}, respectively, and per-sample results are listed in Table~\ref{tab:ziess_overall_viability}.

We also applied UMAP dimensionality reduction \cite{UMAP:20} to individual cell embeddings, computed separately for brightfield and phase images. For the stained samples (85,304 cells from V1 and V2), points are colored by their fluorescence-derived labels. Live and dead cells form clearly separated clusters in the brightfield embedding, and dead cells occupy a distinct region in the phase embedding, although with more overlap. For the unstained samples (199,914 cells), where cell-level ground truth is unavailable, points are colored using predictions from the CNN trained on stained V1 data. Although the separation between predicted live and dead cells is less distinct than in the stained case, predicted dead cells lie closer to one another and farther from predicted live cells in both embeddings, suggesting that the CNN has learned discriminative features that generalize beyond stained samples.

\FloatBarrier
\begin{table*}[t]
\centering
\footnotesize
\setlength{\tabcolsep}{4pt}
\renewcommand{\arraystretch}{1.1}
\caption{Per-experiment cell counts and viability for stained and unstained Zeiss samples (underlying Figure~\ref{fig:ziess_overall_viability_result}). Each stained Zeiss experiment was also measured with PLX to verify that the ground-truth labels on the Zeiss data were assigned correctly. GT labels are unavailable for unstained images (--); NA marks experiments with no stained data.}
\begin{tabular}{@{}l r r rrr rrr rrr@{}}
\toprule
& & & \multicolumn{3}{c}{GT} & \multicolumn{3}{c}{CNN Trained on V1} & \multicolumn{3}{c}{CNN Trained on V2} \\
\cmidrule(lr){4-6} \cmidrule(lr){7-9} \cmidrule(l){10-12}
Experiment & PLX Viab. & Tot.
  & Live & Dead & Viab.
  & Live & Dead & Pred. Viab.
  & Live & Dead & Pred. Viab. \\
\midrule
Day1-V1-Stained            & 96.33\% & 2965  & 2864  & 101  & 96.59\% & 2904& 61& 97.94\% & 2935& 30& 98.99\%\\
Day1-V1-Unstained          & --      & 5204  & --    & --   & --      & 5092& 112& 97.85\% & 5138& 66& 98.73\%\\
\midrule
Day4-V1-Stained            & 94.33\% & 2591  & 2439  & 152  & 94.13\% & 2456  & 135  & 94.79\% & 2468& 123& 95.25\%\\
Day4-V1-Unstained          & --      & 4961  & --    & --   & --      & 4717  & 244  & 95.08\% & 4844& 117& 97.64\%\\
\addlinespace[2pt]
Day4-V2-Stained            & 94.33\% & 1794  & 1672  & 122  & 93.20\% & 1702  & 92   & 94.87\% & 1700& 94& 94.76\%\\
Day4-V2-Unstained          & --      & 3987  & --    & --   & --      & 3775  & 212  & 94.68\% & 3861& 126& 96.84\%\\
\midrule
Day6-V1-Stained            & 94.00\%    & 7008  & 6635  & 373  & 94.68\% & 6600  & 408  & 94.18\% & 6620& 388& 94.46\%\\
Day6-V1-Unstained          & --      & 13774 & --    & --   & --      & 13073 & 701  & 94.91\% & 13215& 559& 95.94\%\\
\addlinespace[2pt]
Day6-V2-Stained            & 94.00\%    & 5022  & 4730  & 292  & 94.19\% & 4695  & 327  & 93.49\% & 4701& 321& 93.61\%\\\
Day6-V2-Unstained          & --      & 10272 & --    & --   & --      & 9621  & 651  & 93.66\% & 9756& 516& 94.98\%\\
\midrule
Day8-V2-Stained& 93.33\% & NA    & NA    & NA   & NA      & NA    & NA   & NA      & NA & NA & NA \\
\addlinespace[2pt]
Day8-V2-Unstained          & --      & 19070 & --    & --   & --      & 17784 & 1286 & 93.26\% & 17983& 1087& 94.30\%\\
\midrule
Day12-V1-Stained           & 75.00\%    & 11738 & 8932  & 2806 & 76.09\% & 8415  & 3323 & 71.69\% & 8455& 3283& 72.03\%\\
Day12-V1-Unstained         & --      & 21837 & --    & --   & --      & 15715 & 6122 & 71.97\% & 17439& 4398& 79.86\%\\
\addlinespace[2pt]
Day12-V2-Stained           & 90.33\% & 12518 & 11121 & 1397 & 88.84\% & 10878 & 1640 & 86.90\% & 10950 & 1568 & 87.47\% \\
Day12-V2-Unstained         & --      & 27598 & --    & --   & --      & 22887 & 4711 & 82.93\% & 24210 & 3388 & 87.72\% \\
\midrule
Day13-V1-Stained           & 69.00\%    & 9187  & 6316  & 2871 & 68.75\% & 6033  & 3154 & 65.67\% & 6215 & 2972 & 67.65\% \\
Day13-V1-Unstained         & --      & 22904 & --    & --   & --      & 15608 & 7296 & 68.15\% & 16673 & 6231 & 72.80\% \\
\addlinespace[2pt]
Day13-V2-Stained           & 87.67\% & 11214 & 9632  & 1582 & 85.89\% & 9398  & 1816 & 83.81\% & 9495 & 1719 & 84.67\% \\
Day13-V2-Unstained         & --      & 25791 & --    & --   & --      & 20939 & 4852 & 81.19\% & 21664 & 4127 & 84.00\% \\
\midrule
Day14-V1-Stained           & 64.50\%  & 9607  & 6427  & 3180 & 66.90\% & 6194  & 3413 & 64.47\% & 7198 & 2409 & 74.92\% \\
Day14-V1-Unstained         & --      & 18633 & --    & --   & --      & 11804 & 6829 & 63.35\% & 14019 & 4614 & 75.24\% \\
\addlinespace[2pt]
Day14-V2-Stained           & 86.33\% & 11660 & 9886  & 1774 & 84.79\% & 9452  & 2208 & 81.06\% & 9615 & 2045 & 82.46\% \\
Day14-V2-Unstained         & --      & 25883 & --    & --   & --      & 20507 & 5376 & 79.23\% & 21661 & 4222 & 83.69\% \\
\bottomrule
\end{tabular}
\label{tab:ziess_overall_viability}
\end{table*}

\end{document}